\documentclass[sn-mathphys-num,iicol]{sn-jnl}% Math and Physical Sciences Numbered Reference Style
\usepackage[T1]{fontenc}
\usepackage{graphicx}
\usepackage{amsmath}
\usepackage{multicol}
\usepackage{multirow}
\usepackage{booktabs}
\usepackage{comment}
\usepackage{array}
\usepackage{booktabs}
\usepackage[T1]{fontenc}
\usepackage{lilyglyphs}
\usepackage{musicography}

\usepackage{siunitx}
\usepackage{svg}
\usepackage{subcaption}
\usepackage{tikz}
\usepackage{listings}
\usepackage{float}
\usepackage{stfloats}
\usepackage{amsfonts}
\usepackage{xcolor}
\usepackage{hyperref}
\usepackage{academicons}

\usepackage{cleveref}%
\usepackage{numprint}

\lstdefinelanguage{LilyPond}{
  keywords={\clef, \time, relative, version},
  keywordstyle=\color{blue}\bfseries,
  comment=[l]\%, % comment style
  commentstyle=\color{gray},
  stringstyle=\color{orange},
  basicstyle=\ttfamily\scriptsize\color{black},
  morestring=[b]",
  alsoletter={\#},
  breaklines=true,
  breakatwhitespace=true,
  frame=0,
  columns=fullflexible,
  xleftmargin=0pt,
  framexleftmargin=0pt,
  aboveskip=0pt,
  belowskip=0pt,
  keepspaces=true,
  showspaces=false,
  showstringspaces=false,
  numberstyle=\tiny\color{gray},
  stepnumber=1,
  breakindent=0pt,
}

\lstdefinelanguage{LilyPondPred}[]{LilyPond}{
  keywordstyle=\color{blue}\bfseries\color[HTML]{3C8031},
  basicstyle=\ttfamily\scriptsize\color[HTML]{3C8031},
}

\crefname{figure}{Fig.}{Figures}
\Crefname{figure}{Fig.}{Figures}
\crefname{section}{Section}{Sections}
\Crefname{section}{Section}{Sections}
\crefname{equation}{Equation}{Equations}
\Crefname{equation}{Equations}{Equation}
\crefname{appendix}{Appendix}{Appendix}
\Crefname{appendix}{Appendix}{Appendix}
\crefname{table}{Table}{Table}
\Crefname{table}{Table}{Table}

\NewDocumentCommand{\musSymbolClef}{ m }{%
\smash{\kern-1pt\musFont\raisebox{1.1ex}{#1}\kern0.3em}%
}

\newcommand{\doiurl}[1]{DOI: \href{https://doi.org/#1}{\nolinkurl{#1}}}

\newcommand{\eqtau}{\stackrel{\epsilon}{=}}

\DeclareRobustCommand{\orcid}[1]{%
  \href{https://orcid.org/#1}{%
    \raisebox{-0.25ex}{\includegraphics[height=2.25ex]{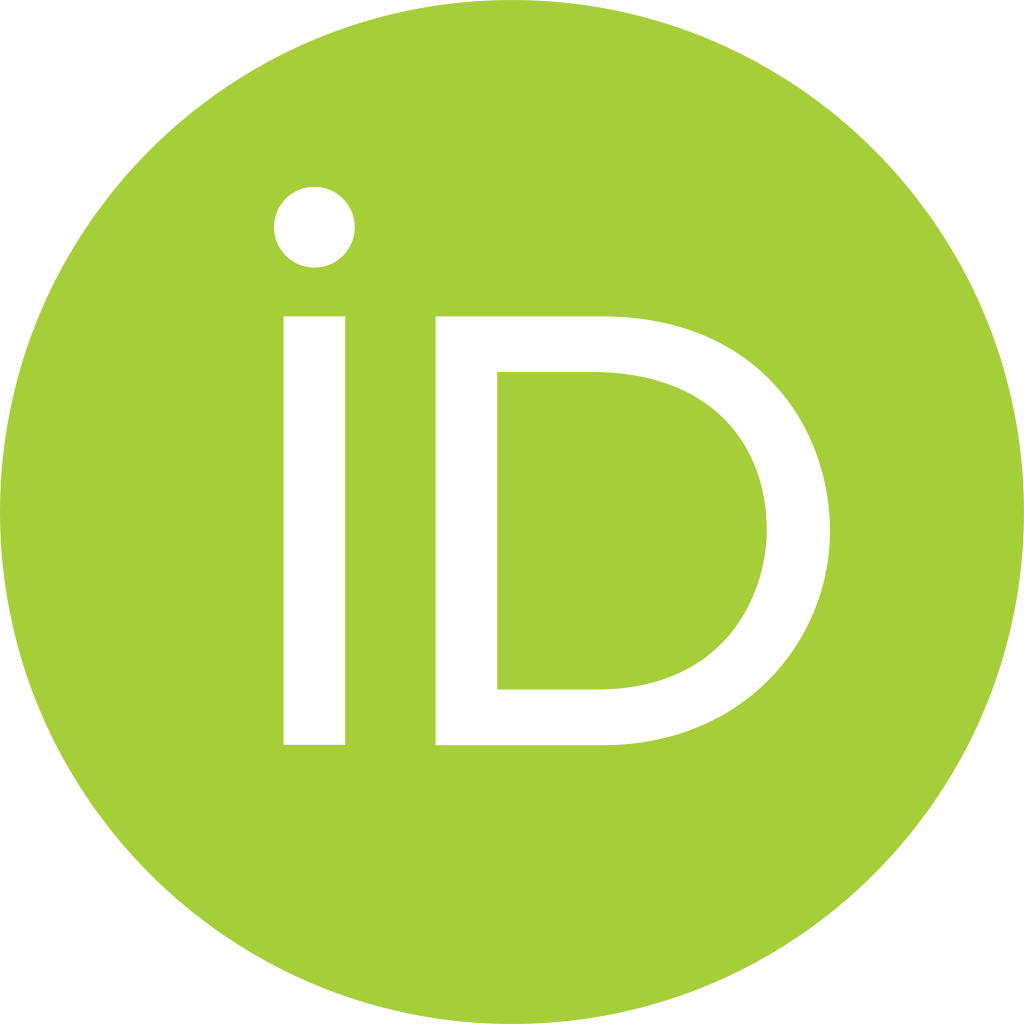}}%
  }%
}

\makeatletter
    \newcommand{\showfont}{
        \begin{itemize}
            \item encoding: \f@encoding{}
            \item family: \f@family{}
            \item series: \f@series{}
            \item shape: \f@shape{}
            \item size: \f@size{}
        \end{itemize}
    }
\makeatother

\begin{document}
\title{A document is worth a structured record: Principled inductive bias design for document recognition}

% \titlerunning{Document recognition as transcription}

%\date{July 2024}

\author*[1,3]{
\orcid{0009-0006-5609-2700}~\fnm{Benjamin} \sur{Meyer}}\email{mebr@zhaw.ch}
% Note: ORCID 0009-0006-5609-2700 was present in original

\author[1,5]{\orcid{0000-0001-5878-8425}~\fnm{Lukas} \sur{Tuggener}}\email{lukas.tuggener@rwai.ch}
% Note: ORCID 0000-0001-5878-8425 was present in original

\author[2]{\orcid{0009-0007-7590-664X}~\fnm{Sascha} \sur{H\"anzi}}\email{haez@zhaw.ch}
% Note: ORCID 0009-0007-7590-664X was present in original

\author[2]{\orcid{0009-0001-8648-2869}~\fnm{Daniel} \sur{Schmid}}\email{scdd@zhaw.ch}
% Note: ORCID 0009-0001-8648-2869 was present in original

\author[1]{\orcid{0009-0004-7783-3964}~\fnm{Erdal} \sur{Ayfer}}\email{ayfererd@students.zhaw.ch} % Assuming ayfr@zhaw.ch from the original commented email list
% Note: ORCID 0009-0004-7783-3964 was present in original

\author[3]{\orcid{0000-0001-8560-2120}~\fnm{Benjamin F.} \sur{Grewe}}\email{bgrewe@ethz.ch}
% Note: ORCID 0000-0001-8560-2120 was present in original

\author[1]{\orcid{0000-0003-4679-8081}~\fnm{Ahmed} \sur{Abdulkadir}}\email{abdk@zhaw.ch}
\equalcont{These authors contributed equally to this work.}
% Note: ORCID 0000-0003-4679-8081 was present in original

\author[1,4]{\orcid{0000-0002-3784-0420}~\fnm{Thilo} \sur{Stadelmann}}\email{stdm@zhaw.ch}
\equalcont{These authors contributed equally to this work.}
% Note: ORCID 0000-0002-3784-0420 was present in original

\affil*[1]{\orgdiv{Centre for AI}, \orgname{Zurich University of Applied Sciences}, \orgaddress{\city{Winterthur}, \country{Switzerland}}}
\affil[2]{\orgdiv{Institute of Product Development and Production Technologies}, \orgname{Zurich University of Applied Sciences}, \orgaddress{\city{Winterthur}, \country{Switzerland}}}
\affil[3]{\orgdiv{Institute of Neuroinformatics}, \orgname{University of Zurich and ETH Zurich}, \orgaddress{\city{Zurich}, \country{Switzerland}}}
\affil[4]{\orgname{European Centre for Living Technology}, \orgaddress{\city{Venice}, \country{Italy}}}
\affil[5]{\orgname{RWAI Schweiz AG}, \orgaddress{\city{Winterthur}, \country{Switzerland}}}

% 4 to 6 keywords

%\twocolumn[
%\begin{@twocolumnfalse}

%
\maketitle              % typeset the header of the contribution

\begin{abstract}{}
Many document types use intrinsic, convention-driven structures that serve to encode precise and structured information, such as the conventions governing engineering drawings. However, many state-of-the-art approaches treat document recognition as a mere computer vision problem, neglecting these underlying document-type-specific structural properties, making them dependent on sub-optimal heuristic post-processing and rendering many less frequent or more complicated document types inaccessible to modern document recognition. We suggest a novel perspective that frames document recognition as a transcription task from a document to a record. This implies a natural grouping of documents based on the intrinsic structure inherent in their transcription, where related document types can be treated (and learned) similarly. We propose a method to design structure-specific relational inductive biases for the underlying machine-learned end-to-end document recognition systems, and a respective base transformer architecture that we successfully adapt to different structures. We demonstrate the effectiveness of the so-found inductive biases in extensive experiments with progressively complex record structures from monophonic sheet music, shape drawings, and simplified engineering drawings. By integrating an inductive bias for unrestricted graph structures, we train the first-ever successful end-to-end model to transcribe mechanical engineering drawings to their inherently interlinked information. Our approach is relevant to inform the design of document recognition systems for document types that are less well understood than standard OCR, OMR, etc., and serves as a guide to unify the design of future document foundation models.
\end{abstract}

%\end{@twocolumnfalse}
%]

%\clearpage

\begin{figure*}[!t]
\centering
\includegraphics[width=\textwidth,keepaspectratio,valign=m,margin=0.02cm]{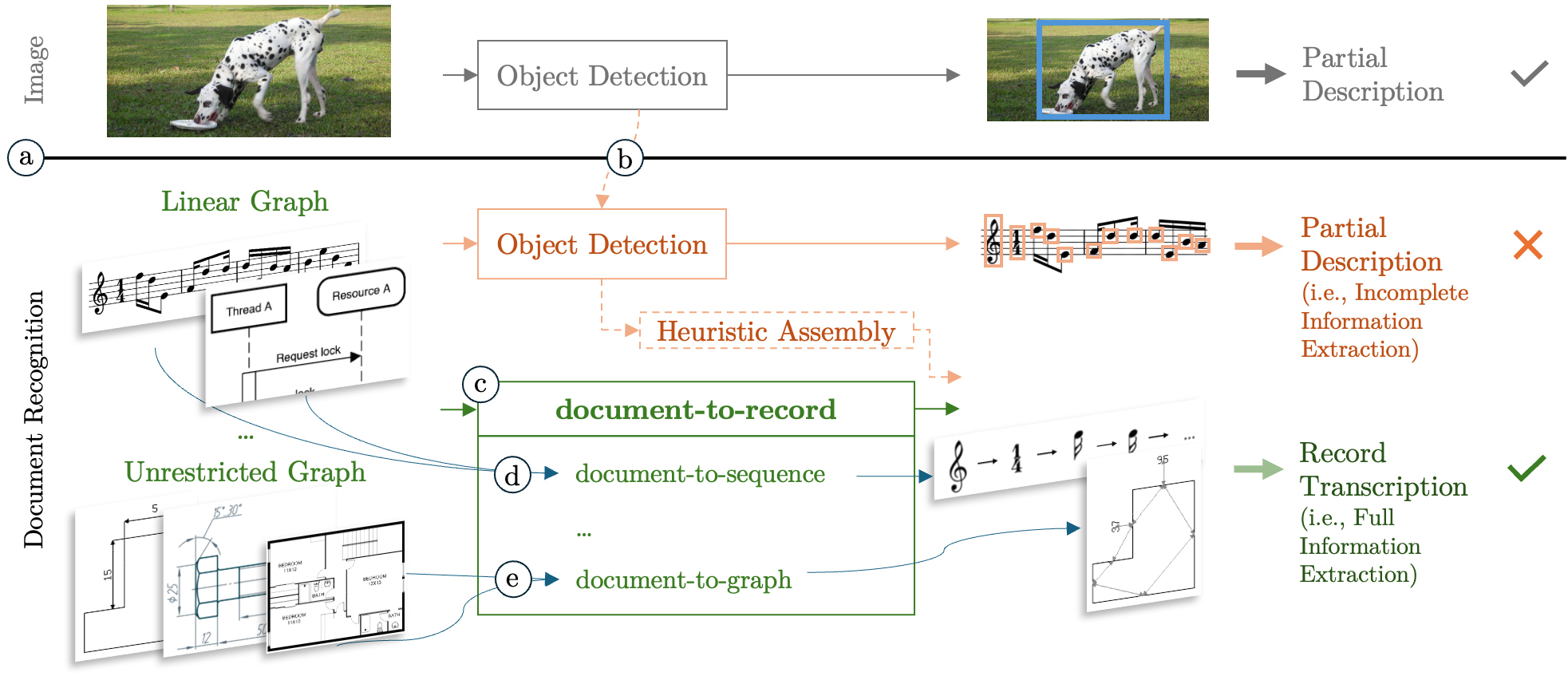}
\caption{
Document recognition and natural image analysis are often approached with similar methods, however:
(a) Image analysis is about extracting partial descriptions of complex and ambiguous content (e.g., detecting the ``dog'' object doesn't account for the shadows/dog breed/\dots~in the image). 
In contrast, document recognition is about the complete transcription of explicitly encoded information.
(b) Viewing document recognition merely as a computer vision task results in incomplete extraction, failing to fully capture all the information a document is intended to convey.
% For example, in engineering drawings, mapping tolerance annotations to shape outlines requires visual clues like arrows and is complicated to implement algorithmically. 
(c) We propose the \emph{document-to-record transcription} task, formulating document recognition as the extraction of complete information (i.e., a record).
The record structure is specific to each document type, but archetypes of record structures exist that span multiple document types. Thus formed document classes give rise to a principled way of designing inductive biases for the underlying learning systems, e.g.:
(d) Notes in monophonic music or exchanged messages in a sequence diagram form linear graphs, implying the use of a structure-specific inductive bias for \emph{document-to-sequence} transcription; (e) more complex record types, such as for engineering drawings or floor plans, form unrestricted graphs and require \emph{document-to-graph} transcription.  
% , establishing an end-to-end learning framework for document understanding. 
% Furthermore, the domain-intrinsic record structure links document types on a principled level, such as documents with a sequential record (document-to-sequence) or a graphical record (document-to-graph).
} \label{fig:abstract}
\end{figure*}

\section{Introduction} \label{sec:introduction}

Deep learning-based computer vision systems have become increasingly powerful in processing natural images \cite{stadelmann2018deep,chai2021deep}, which has led to their application in various types of digital document recognition tasks that process pixel-based representations of said documents~\cite{subramani2020survey}.
While this adoption has proven successful \cite{rios-vila_sheet_2024,meier_fully_2017,li_trocr_2023,schmitt-koopmann_mathnet_2024,wei2024general}, there are many highly relevant document types for which success remains limited, such as engineering drawings \cite{sarkar_automatic_2022}, floor plans \cite{rezvanifar_symbol_2020}, and others \cite{uzair_electronet_2023,mardiana2024comparative,gada_object_2021}, despite such documents being easily understood by domain experts.
We postulate that overcoming existing performance gaps requires a shift in perspective on automated document analysis that formulates the analysis task differently than an isolated computer vision task (e.g., object detection), and instead as transcription (cp. \cref{fig:abstract}).
This new perspective implies that specific inductive biases can be designed that reflect the intrinsic structures found in broad classes of documents.
The remainder of this section will establish this perspective, starting from the status quo.

Many of the methods employed in both natural image analysis and document recognition were first developed for natural image datasets \cite{russakovsky2015imagenet,minaee2021image,zou2023object} and subsequently adapted for use in document analysis tasks (e.g., \cite{tuggener_deep_2018,yamasaki_apartment_2018}).
Those methods, by design, align with natural images by extracting \emph{partial descriptions}, necessitated by the granular, rich, and dense image content: While extracting \emph{everything} contained in an image is usually infeasible and never attempted, different descriptions enable different downstream tasks.
For example, object detection may identify and locate a set of predefined common objects, making downstream tasks possible such as object counting \cite{buzzy2020real}, object tracking \cite{pal2021deep}, or robotic grasping \cite{khor2024robotic} (\cref{fig:abstract} (a)), but might not suffice to enable image localization \cite{hays2008im2gps}.
However, those partial-description methods do not naturally align with documents.
Documents encode specific (and comparatively little) information that they were explicitly designed to convey,
and analyzing documents is about \emph{precisely extracting} this information \emph{fully}.
Enforcing a partial description, e.g., through object detection to recognize symbols \cite{rezvanifar_symbol_2020}, results in incomplete information extraction, as, for example, the essential relationships between symbols are ignored by a pure object detection method (\cref{fig:abstract} (b)). Of course, such a partial extraction can be (and in practice usually is) amended by individual, heuristic post-processing to recover, e.g., said object relationships. But a more principled, document-centric approach seems possible. 

For this, we first introduce necessary terminology (cf. \cref{sec:formalization} for formal definitions): We refer to the information encoded in a document as its \emph{record}.
This record is both the source and the transcription of a document, includes all information, and thus forms a natural interface to execute downstream tasks.
% The record does not contain visual elements, such as stylistic choices (e.g., line thickness, color, font) and non-semantic layout options (e.g., line spacing, symbol spacing), which are by-products of visually representing the record.
It does not contain visual attributes resulting from visually representing the record as a document, such as stylistic choices (e.g., line thickness, color, font) and non-semantic layout options (e.g., line or symbol spacing).
This record-document duality gives rise to viewing document understanding as an end-to-end, \emph{document-to-record transcription} task (\cref{fig:abstract} (c)).

\begin{figure*}[tb]
\centering
\begin{subfigure}{0.32\textwidth}
\centering
\includegraphics[width=\textwidth,height=65px,keepaspectratio,valign=m,margin=0.02cm]{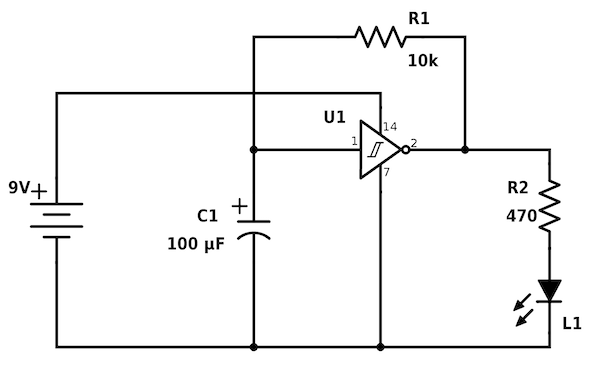}
\caption{
\textit{Electric circuit diagrams:} 
A graph of electronic symbols (e.g., diodes, resistors) with properties (e.g., resistance value). 
% Visual attributes: placement of symbols and wires.
% A record is a general graph of electronic symbols, e.g., earth ground, diode, and resistor, with additional properties, e.g., the resistance value. 
% Solely visual attributes include the non-semantic placement of symbols, optional symbols indicating line crossings, wire spacing, and stylistic options, e.g., line thickness.
}
\end{subfigure}
\hfill
\begin{subfigure}{0.32\textwidth}
\centering
% \includegraphics[width=\textwidth,height=65px,keepaspectratio,valign=m,margin=0.02cm]{img/examples/mathematical_expression_border.png}
% \caption{PLACEHOLDER for document.}
\includegraphics[width=\textwidth,height=65px,keepaspectratio,valign=m,margin=0.02cm]{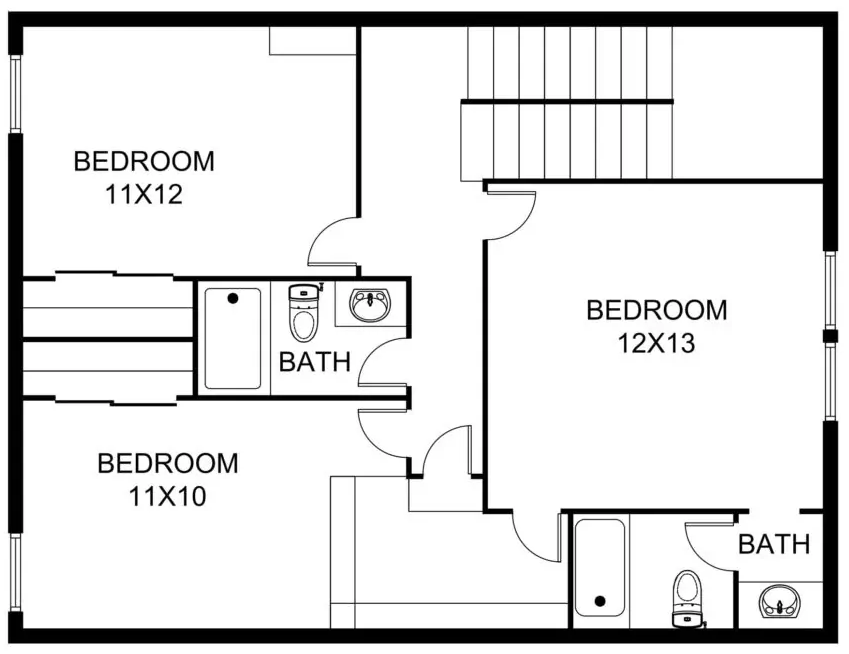}
\caption{
\textit{Floor plans:} 
A graph of shapes (e.g., walls, doors, windows, stairs, fixtures) with lengths to form and dimensionalize rooms.
% Visual attributes: font, line thickness.
% A record is a general graph of primitives, e.g., outside walls, inside walls, doors, and windows with relative lengths that form dimensionalized rooms, as well as stairs, built-in closets, and fixtures, e.g., sinks.
% Solely visual attributes are stylistic options, e.g., font, line thickness, and colors.
}
\end{subfigure}
\hfill
\begin{subfigure}{0.32\textwidth}
\centering
\includegraphics[width=\textwidth,height=65px,keepaspectratio,valign=m,margin=0.02cm]{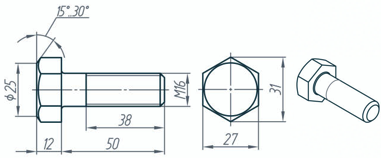}
\caption{
\textit{Engineering drawings:} 
A graph of shapes (e.g., lines) and manufacturing information (e.g., dimensions, tolerances).
% Visual attributes: annotation placement, arrows, help-lines.
% A record is a general graph of objects, including 3D shape primitives, e.g., 3D lines, and product and manufacturing information (PMI), e.g., geometric dimensions, tolerances, and characteristics.
% Solely visual attributes include non-semantic annotation placement, optional annotations, and stylistic options, e.g., the font, line thickness, and arrow styles.
}
\end{subfigure}
\hfill
\begin{subfigure}{0.32\textwidth}
\centering
\includegraphics[width=\textwidth,height=65px,keepaspectratio,valign=m,margin=0.02cm]{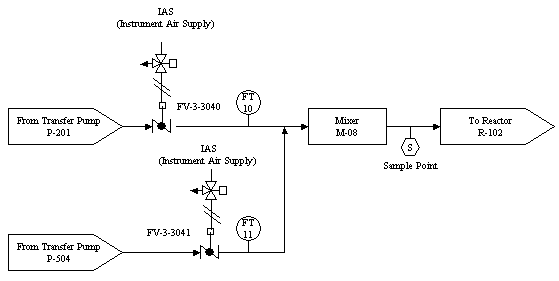}
\caption{
% \textit{Piping and instrumentation diagrams:} 
\textit{P\&ID diagrams:} 
A directed graph of primitives (e.g., valves, tanks, pumps) with properties (e.g., numerator, control, device). %, connected through pipework or electrical connections.
% Visual attributes: non-semantic placement of symbols and wiring.
}
\end{subfigure}
\hfill
\begin{subfigure}{0.32\textwidth}
\centering
\includegraphics[width=\textwidth,height=65px,keepaspectratio,valign=m,margin=0.02cm]{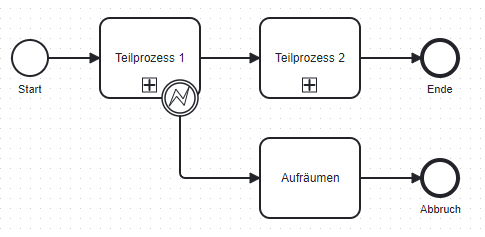}
\caption{
\textit{Business Process Models and Notation:} 
A directed graph of business process elements (e.g., activities, events, gateways).
% Visual attributes: non-semantic spacing.
% Conventions include standardized symbols and connection types \cite{object2011business}.
}
\end{subfigure}
\hfill
\begin{subfigure}{0.32\textwidth}
\centering
% \includegraphics[width=\textwidth,keepaspectratio,valign=m,margin=0.02cm]{img/examples/parking.jpg}
% \caption{{\bf \textcolor{red}{REVAMPING this whole figure. This is very important, but not effective yet. Please IGNORE for now.}}}
\includegraphics[width=\textwidth,height=65px,keepaspectratio,valign=m,margin=0.02cm]{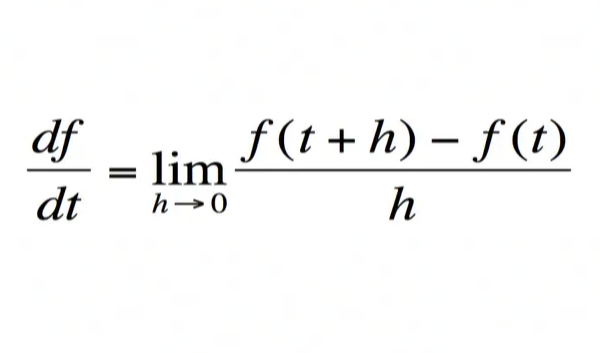}
\caption{
\textit{Mathematical expressions:} 
An ordered tree of mathematical primitives (e.g., fractions, variables, and numbers).
% Visual attributes: font, symbol spacing.
% Convention \cite{ISO80000-2}: standardized symbols.
}
\end{subfigure}
\hfill
\begin{subfigure}{0.32\textwidth}
\centering
% \includegraphics[width=\textwidth,height=65px,keepaspectratio,valign=m,margin=0.02cm]{img/examples/mathematical_expression_border.png}
% \caption{PLACEHOLDER for document.}
\includegraphics[width=\textwidth,height=65px,keepaspectratio,valign=m,margin=0.02cm]{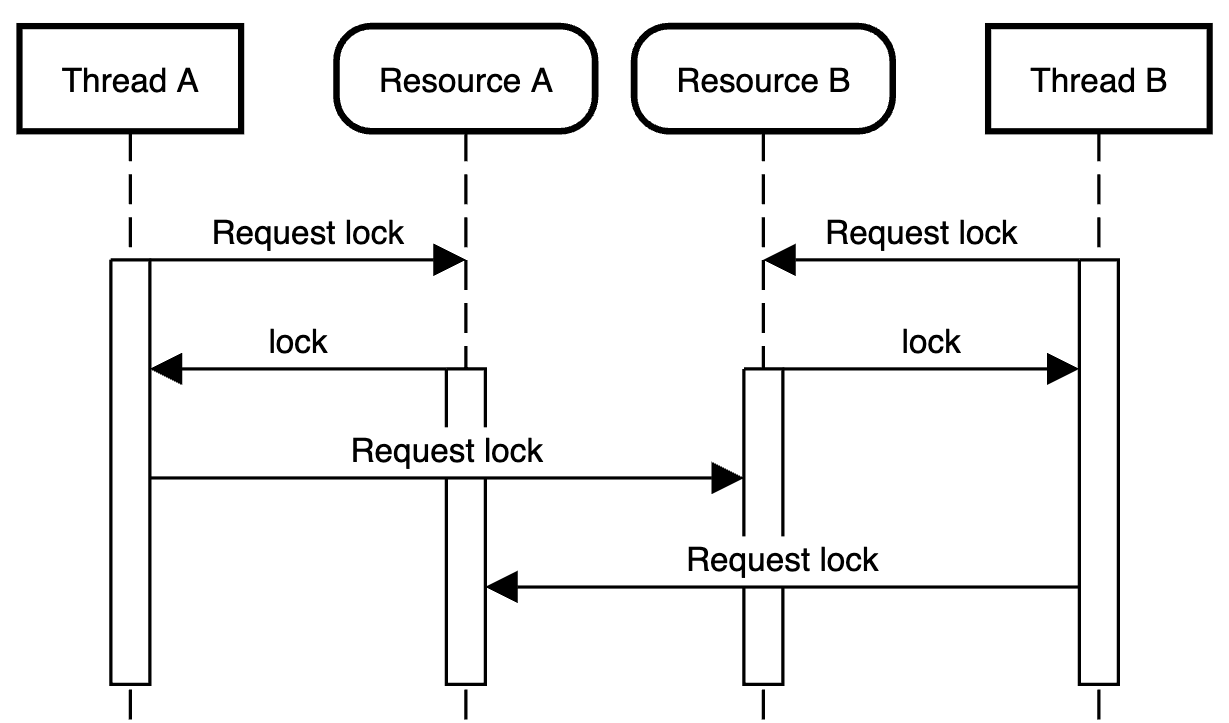}
\caption{
\textit{Sequence diagrams:} 
A sequence of text messages with a source and target entity.
% Visual attributes: non-semantic entity order, entity spacing.
% Convention \cite{OMG:UML:2.5.1}: top-to-bottom ordering, semantic arrow types.  % \cite[chapter 17.8]{OMG:UML:2.5.1}
}
\end{subfigure}
\hfill
\begin{subfigure}{0.32\textwidth}
\centering
\includegraphics[width=\textwidth,height=65px,keepaspectratio,valign=m,margin=0.02cm]{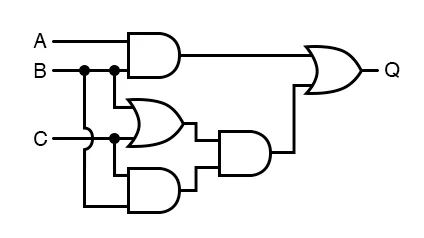}
\caption{
\textit{Logic circuit diagrams:} 
A graph of logic gates (e.g., \texttt{AND} gate, \texttt{OR} gate) connected by wiring.
% Visual attributes: gate placement, non-semantic wire crossing symbol.
% Convention \cite{ANSI-IEEE-91a-1991}: standardized symbols.
}
\end{subfigure}
\hfill
\begin{subfigure}{0.32\textwidth}
\centering
\includegraphics[width=\textwidth,height=65px,keepaspectratio,valign=m,margin=0.02cm]{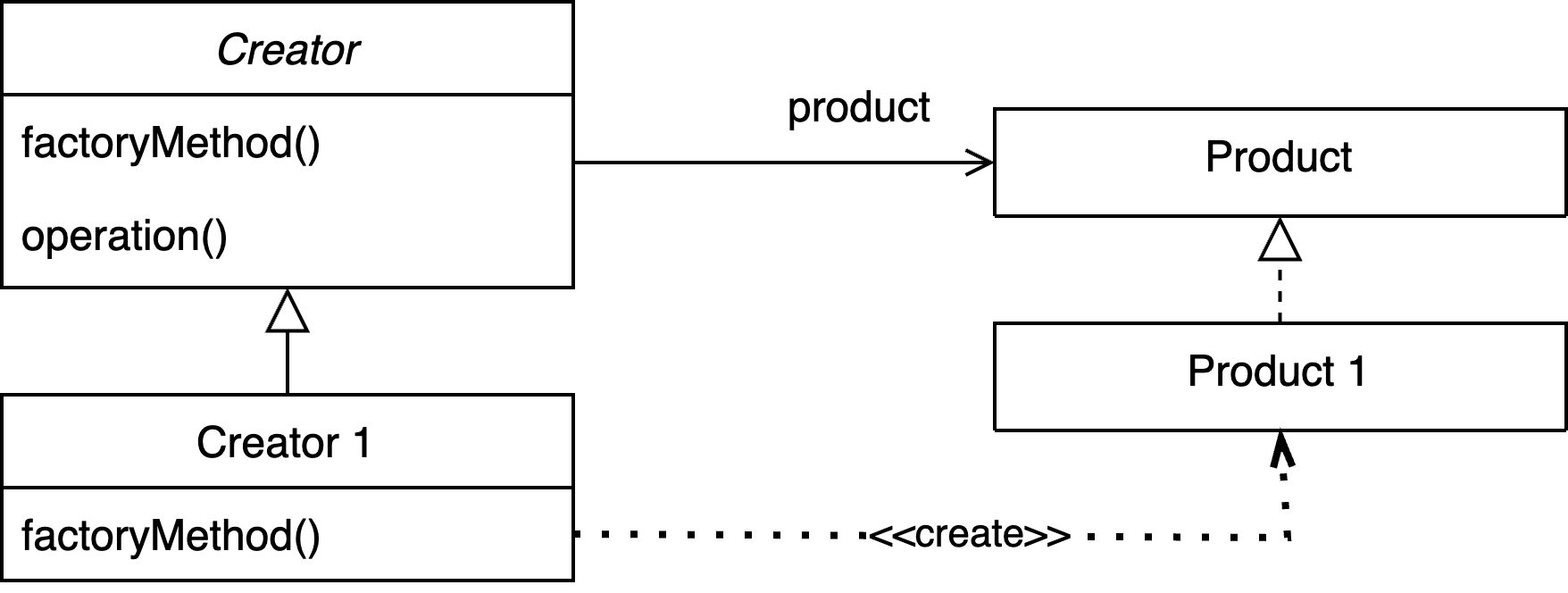}
\caption{
\textit{Class diagrams:} 
A graph of classes with, e.g., names, methods, and relationships (e.g., inheritance).
% Visual attributes: non-semantic class placement.
% Convention \cite{OMG:UML:2.5.1}: semantic arrow types.
}
\end{subfigure}
\caption{
Examples of convention-bound documents with brief descriptions of the contained information (record) and its structure (record structure).
Document types with similar record structures can be analysed with a unified document recognition system under the perspective of our transcription framework.
}
\label{fig:convention-bound-documents}
\end{figure*}

The exact schema of a record is document-type-specific, but, in general, it contains interlinked data, most generally represented by a property graph.
We refer to the nodes in that graph as \emph{record nodes}, which have a node type and type-specific properties.
Record nodes find their correspondence in the respective document in primitive shapes and symbols. 
Relationships (edges) between nodes are typically visualized by clues such as intelligible spacing and additional primitives like arrows (cf. \cref{fig:convention-bound-documents}).

It is interesting to consider the overall structure of the record (henceforth called the \emph{record structure}):
%This notion of a record and its specific structure implies that certain document types will share a common, intrinsic high-level record structure,  
% that is, intrinsic structures in the information they exchange, 
%naturally grouping otherwise unrelated document types by these structural cues:
For example, notes in monophonic sheet music form a linear graph (or: a sequence) where nodes 
are connected in a staff, having only one predecessor and one successor
% have neighbors 
-- a sequential record structure, as with standard text.
Record nodes in the record of mathematical expressions form ordered trees, in which nodes may have nested child nodes, but children form an ordered sequence -- a hierarchical sequential structure.
Document types with less rigid content order form general graphs where nodes have interlinked relations -- engineering drawings fall into this category, containing primitive shapes and annotations that, through visual cues, are interlinked in complex ways to form a discrete $3$-dimensional shape with manufacturing annotations. Floor plans are another example.

Interestingly, record structures thus naturally group otherwise unrelated document types together, e.g., monophonic sheet music and plain text by means of their sequential record structure. This warrants the specialization of document-to-record into tasks such as document-to-sequence, document-to-set, and document-to-graph (others exist that haven't been exemplified above, e.g., document-to-tree for diagrams like mind maps or organizational charts). Each possible specialization makes different assumptions about the restrictions present in the record structure, mapping to known types of graphs from graph theory \cite[Chapter 2.3]{10.5555/22577} (\cref{fig:abstract} (d) and (e)).

This structural grouping explains the performance gap in document recognition between certain document types: 
Recent successes in end-to-end methods are limited to documents with an inherently sequential record structure, such as in OCR (Optical Character Recognition, e.g, \cite{li_trocr_2023}) and OMR (Optical Music Recognition, which is sequential in the sequence of notes when processing polyphony and multiple staves in a predefined order, e.g.~\cite{rios-vila_sheet_2024}).
These methods address the document-to-sequence problem by adopting sequence-to-sequence architectures designed with a sequential bias in their next-token generation~\cite{sutskever2014sequence}.
% Similar works \cite{schmitt-koopmann_mathnet_2024,lee_pix2struct_2023} follow a related approach, but technically without a \emph{uniform} record representation in their training targets, yet emphasizing the usefulness of transcription by seeking uniformity through normalization to aid learning.
In contrast, most document types currently lacking complete information extraction solutions carry an inherently non-sequential, more general record structure, requiring the extraction of a graph structure, where common methods lack adequate inductive biases.
% In summary, \emph{records contain intrinsic structures and models must use adequate structural biases.}

This analysis also highlights a blind spot in attempts to build foundation models for documents such as those in \cite{wei2024general,poznanski2025olmocr,wang2024qwen2,hamdi2025vista}: Present approaches exclusively employ sequence-to-sequence methods and hence focus on document types that carry a mostly sequential record structure.
Our perspective offers a path to incorporating suitable inductive biases for non-sequential document types into such models 
% and highlights directions to improve general performance 
by adapting them to more general structures or by motivating the normalization of training targets to align with current inductive biases, e.g., through heuristic graph linearization strategies~\cite{xypolopoulos2024graph}.
% canonical forms \cite{schmitt-koopmann_mathnet_2024} or

Strictly speaking, the above does not hold for all types of documents (e.g., not for comic books or line art). In this work, we instead focus on documents designed for \emph{precise} information exchange (see \cref{fig:convention-bound-documents} for examples).
We term such documents \emph{convention-bound} as they follow strict conventions that govern the contained information and emerged to ensure unequivocal information exchange among domain experts (see \cref{sec:formalization} for a formal definition of convention-bound documents).
These include sheet music or engineering drawings, but exclude imprecise documents like (ambiguous) illustrative diagrams.
Due to their non-ambiguity, the document-to-record framing is well-posed for convention-bound documents.

From this consideration, we derive our main conceptual advance: 
Document recognition on convention-bound documents is best seen as document-to-record transcription, and benefits %-- next to properties that make documents a special class of images -- 
from incorporating the intrinsic record structure as an inductive bias in the model's architecture and training process. This enables end-to-end learning for more rare and less-understood document types, particularly those with a non-sequential nature.
% Document recognition on convention-bound documents is document-to-record transcription, and effective transcription necessitates incorporating the intrinsic record structure present in document types as an inductive bias in the model's architecture and training process.

% Contributions
In this paper, our contributions are: 
(1)~\emph{``Document recognition as transcription'' perspective:} By viewing document recognition as complete record transcription of convention-bound documents rather than computer-vision-based extraction of partial descriptions, a meaningful and efficient grouping of document types by record structure emerges naturally that explains performance gaps in state-of-the-art approaches; it is later leveraged for learning.
%This frames document recognition and understanding as an (end-to-end) document-to-record task;
%(2)~\emph{convention-bound documents:} Defining subsets of convention-bound document types, enabling a meaningful and efficient grouping of document types by their natural record structure that is leveraged for learning;
% (3)~\emph{Record structure as bias:} Identifying principles to incorporate specific inductive biases that cover all types of convention-bound documents;
%(3)~\emph{Record structure as inductive bias:} Highlighting different domain-intrinsic record structures and their impact on solving tasks; and
(2)~\emph{Method for systematically designing inductive biases for document recognition systems:} Utilizing said record structures in a principled, systematic way leads to a novel approach in designing document-centric model architectures applicable across document types. This establishes a bridge from document-to-record transcription to methods for graph prediction.
(3)~\emph{Implementation of a practical respective machine learning framework:} We introduce an end-to-end, domain-agnostic learning framework based on a unified transformer backbone with document-group-specific adapted inductive biases, useful for sample-efficient document foundation models.
(4)~\emph{Comprehensive experimental evaluation of proofs of concept:} The postulated principles are valid and effective in monophonic sheet music (document-to-sequence), shape drawings (document-to-set), and simplified engineering drawings (document-to-graph). 
For mechanical engineering drawings, we show the first ever working implementation of an end-to-end learned document recognition approach.

\section{Related Work}

\subsection{Information extraction from convention-bound documents}

Document recognition of convention-bound documents can be viewed as information extraction.
Solutions include:

\textit{Incomplete information extraction.} Computer vision methods such as object detection (symbol spotting) and semantic segmentation have been widely applied to convention-bound documents, including newspaper pages \cite{meier_fully_2017}, sheet music \cite{hajic_towards_2018,tuggener_deep_2018,tuggener2021deepscoresv2}, mathematical expressions \cite{schmitt2022formulanet}, engineering drawings \cite{sarkar_automatic_2022}, circuit diagrams \cite{uzair_electronet_2023}, floor plans \cite{rezvanifar_symbol_2020,kim_automatic_2021,seo2020inference}, UML diagrams \cite{mardiana2024comparative}, and P\&ID diagrams \cite{gada_object_2021}. However, these approaches are inherently incomplete as they typically detect individual symbols or regions but fail to capture the semantic relationships or structure between symbols directly, which is essential for fully understanding and utilizing such documents.

\textit{Heuristic full information extraction.} 
Full information extraction (i.e., transcription) involves not just extracting relevant symbols and shapes but also their corresponding relationships.
To achieve this, incomplete information extraction methods have been extended by post-hoc heuristic symbol assembly steps, for example, in UML diagrams \cite{huber2020work}, or sheet music \cite{hajic_towards_2018}.

\textit{Narrow end-to-end information extraction.} Many methods employ sequence-to-sequence approaches to learn robust record extraction end-to-end for \emph{single} types of documents.
These records follow either an explicit sequential structure, as in text recognition \cite{li_trocr_2023}, or an implicit sequential structure inherent to ordered tree structures through their universal address system \cite[chapter 11.3]{rosen_discrete_2013}, as in mathematical expressions \cite{schmitt-koopmann_mathnet_2024} and visual language understanding \cite{lee_pix2struct_2023}. 
The authors of \cite{rios-vila_sheet_2024} apply end-to-end information extraction to general sheet music, where their record representation follows an (approximately) sequential structure.
Moving beyond sequential biases, a few recent approaches have begun to apply image-to-graph architectures \cite{cong2023reltr,shit2022relationformer} to directly extract nonlinear relational structures from technical drawings, such as general graphs from piping and instrumentation diagrams (P\&IDs) \cite{sturmer2025engineering} and simplified geometric graphs from floor plans \cite{hu2024raster}.
% as \texttt{**kern} code 
% Historic end-to-end optical character recognition systems follow a specialized architecture abusing the sequential structure in some form \cite{lecun_gradient-based_1998}, whereas recent approaches applying sequence-to-sequence architectures \cite{}.
% The sequential record is represented using a common code format, which may not guarantee a uniform record representation. For example, LaTeX can express the same formula in various forms of code. In line with our analysis, normalization attempts that aim to achieve more uniform representations improve performance \cite{schmitt-koopmann_mathnet_2024,lee_pix2struct_2023}, emphasizing the usefulness of precise transcription.

\textit{Broad end-to-end information extraction.}
Recent works also include early attempts at building document foundation models \cite{zhang2024document}, which can extract text, lists, tables, equations, music notation, and charts from entire documents \cite{wei2024general,poznanski2025olmocr,wang2024qwen2,hamdi2025vista} or parts of documents \cite{blecher2023nougat,hu2024mplug}.
Those models follow the natural reading order, assuming an approximately sequential record structure, explaining the success of adapting sequence-to-sequence methods to respective document types. Scaling such models to less sequential document types would require either massive amounts of training data (with massive compute due to a noisy training signal) to learn the respective structure from the data alone \cite{stadelmann2022data, luley2023concept}, or else need to be equipped with appropriate but still broad inductive biases to be efficient \cite{tuggener2024so}.
 
%\smallskip
Our classification of document types according to their intrinsic record structure reveals a research bias towards documents with inherently sequential organization in the current literature, where document-to-sequence methods are commonly used.
In contrast, complete information extraction remains lacking for inherently non-sequential document types.
By shifting the research perspective and introducing a principled approach for developing more comprehensive models, supported by the experimental evidence presented below, we aim to address the uncharted territory left by recent methods, enabling data-efficient document foundation models for a broader array of document types.

\subsection{Relational inductive biases} \label{sec:related-work:relational-bias}

The concept of relational inductive biases \cite{battaglia2018relational} provides a framework for understanding how neural architectures exploit dependencies in data. 
Historically, this perspective was driven by intuitive architectural designs tailored to inherent data structures (e.g, CNNs for grids, DeepSets for sets).
More recently, these architectural biases have been formalized, notably through group theory for symmetric relations \cite{bronstein2017geometric} and category theory for a broader mathematical foundation \cite{gavranovic2024position,shiebler2021category}.
While these frameworks primarily focus on the processing of relational \textit{inputs}, document-to-record transcription extends this challenge to predicting structured (relational) \textit{outputs} from rasterized images (in which relations are conveyed visually). %Nevertheless, from a relational viewpoint, mapping from documents to these explicit structures still motivates the incorporation of appropriate inductive biases.
As the concern for appropriate relational inductive biases is a shared one, we now review related literature and position the proposed sequence, set, and graph prediction methods within this broader research landscape.

\textit{Graph generation.} 
A common strategy for learning graph distributions is to linearize graphs into flat sequences.
However, because arbitrary graphs lack a natural sequential order, this process yields up to $(n+m)!$ valid permutations for a single graph with $n$ nodes and $m$ edges. Naively training on these permutations forces the model to learn a highly diluted distribution.
To mitigate this exponential explosion, methods typically restrict the permutation space using heuristics \cite{you2018graphrnn,chen2025graph} or by learning custom linearization strategies \cite{wang2025learning}.
Still, such heuristics must be applied carefully, as sequence ordering impacts learning dynamics \cite{vinyals2015order}.
Alternatively, diffusion-based methods avoid imposing sequential structures by iteratively editing noisy graphs to match a target distribution—an approach that enables the use of permutation-equivariant architectures \cite{vignac2022digress}.

In contrast to graph generation, document transcription is a deterministic, many-to-one graph prediction task, as will be formalized in \cref{eq:convention-bound}. This bypasses any need to learn distributions over graphs: While many-to-one prediction can be framed and potentially learned as a conditional Dirac target distribution, we suspect the complex distributional modeling used in graph generation is unhelpful for document transcription.
Instead, our graph prediction bias design will incorporate teacher forcing based on a node-first linearization, yielding a permutation space of $n!m!$ valid sequences (see Section \ref{sec:base-architecture}).
The results will show that despite this high variance in the decoder inputs during training, the decoder learns the transcription task by successfully following its remaining-node order during inference (see Section \ref{sec:results-appropriate}), and  
future work will investigate if more strictly enforcing the model's dynamically learned remaining-node order during training leads to more robust transcription capabilities.

\textit{Sequence prediction.} 
Left-to-right autoregressive sequence prediction \cite{sutskever2014sequence} has achieved immense success \cite{radford2019language,brown2020language}, particularly via causal self-attention.
Building on this foundation, we adapt a standard causal (sequence) transformer to support more diverse structural biases through targeted modifications to only the attention mask and loss function. 

Our proposed sequence prediction aligns very well with the literature (see Section \ref{sec:seq-to-seq}). 

\textit{Set prediction.} 
DETR \cite{carion2020end} reformulates object detection as a set prediction problem, implemented using learnable ``object queries'' and a set loss based on bipartite matching between predictions and ground truth.
While the methodology is applicable to general set prediction, DETR and many subsequent works \cite{zhu2020deformable,li2022dn,zhang2022dino} focus on object detection and use domain-specific terminology (e.g., ``object queries'' rather than general ``element queries''). To align with this established literature, we adopt this terminology here.
Recent related approaches include Slot Attention \cite{locatello2020object} and Set Transformer \cite{lee2019set}; both use a form of cross-attention to bottleneck a set of inputs into a pooled set of representations, referred to as ``slots'' or ``seed vectors'', that are functionally analogous to learned object queries.
% Slot Attention uses "slots"—a set of representations that compete across the input features and bottleneck them—instead of learnable object queries.
% Similarly, the Set Transformer uses "seed vectors" to bottleneck a set of inputs into a pooled set of features, e.g., to learn a set of clusters.

Learnable object queries have two known issues:
First, they converge slowly due to the instability of bipartite graph matching in early training stages \cite{li2022dn}. Second, they require a number of queries significantly larger than the maximum number of objects in an image, incurring substantial computational overhead during both training and inference \cite{carion2020end}. For example, DINO \cite{zhang2022dino} uses $900$ queries on the COCO dataset, which contains only up to $80$ objects per image.
Recent works have attempted to address these limitations. Convergence instability has been mitigated by injecting noisy ground truth into the object queries \cite{li2022dn}, a technique that functions as a surrogate for teacher forcing by providing the model with ground-truth hints that accelerate convergence. Additionally, dynamic queries have been introduced to prune the set of object queries throughout the forward pass, reducing the computational footprint of deeper decoder layers \cite{hong2025layer}.

While these are working improvements, we propose a more fundamental rethinking of set prediction by replacing bipartite matching entirely with autoregressive remaining-node prediction.
Exhibiting architectural similarity to XLNet \cite{yang2019xlnet}, our approach will natively enable teacher forcing (see Section \ref{sec:seq-to-set}). Furthermore, this formulation will require exactly as many prediction queries as there are objects per sample, providing a strictly minimal query count.

\textit{Graph prediction.} 
Several works extend the set prediction paradigm to graph prediction by incorporating a relation prediction module. Much of this research focuses on the domain of scene graph generation \cite{lu2021context,kim2021hotr,cong2023reltr,shit2022relationformer,li2022sgtr,im2024egtr}, embedding domain-specific terminology (e.g., framing graphs as ``relations'' or ``interactions'' between ``subjects'' and ``objects''). To align with this established literature, we adopt this terminology here.
HOTR \cite{kim2021hotr} utilizes object and interaction queries as two parallel set predictions, deriving multi-label interactions from similarity scores between object representations and projected interaction representations.
RelTR \cite{cong2023reltr} detects subjects and objects using shared entity representations, which are then used to predict directed relations.
Relationformer \cite{shit2022relationformer} introduces a distinct relationship query encoded together with the learnable object queries. For each contextualized object query pair and the contextualized relationship query, a relationship (or absence thereof) is predicted independently. This method has been applied to document transcription \cite{sturmer2025engineering}.

In contrast, our graph prediction method simply models the graph as a set of nodes and edges, reducing graph prediction to a set prediction. Compared to existing methods that predict relations independently for each object pair, our autoregressive design automatically enables the model to account for the emerging graph topology during prediction.

\section{Methods} \label{sec:methods}

\subsection{Overview}
We first formalize the concept of convention-bound documents in \cref{sec:formalization} 
and establish that document-to-record transcription is well-posed for this class of documents.
We then introduce a general learning framework for document-to-record transcription in \cref{sec:framework}, accompanied by a foundational architectural design in \cref{sec:base-architecture}.
Subsequently, we specialize this architecture for document-to-sequence, document-to-set, and, more generally, document-to-graph tasks in \crefrange{sec:seq-to-seq}{sec:seq-to-graph}; each of which is exemplified with an example use case in the next section.

\subsection{Convention-bound documents} \label{sec:formalization}

\begin{figure}[b]
    \centering
    \includegraphics[width=\linewidth]{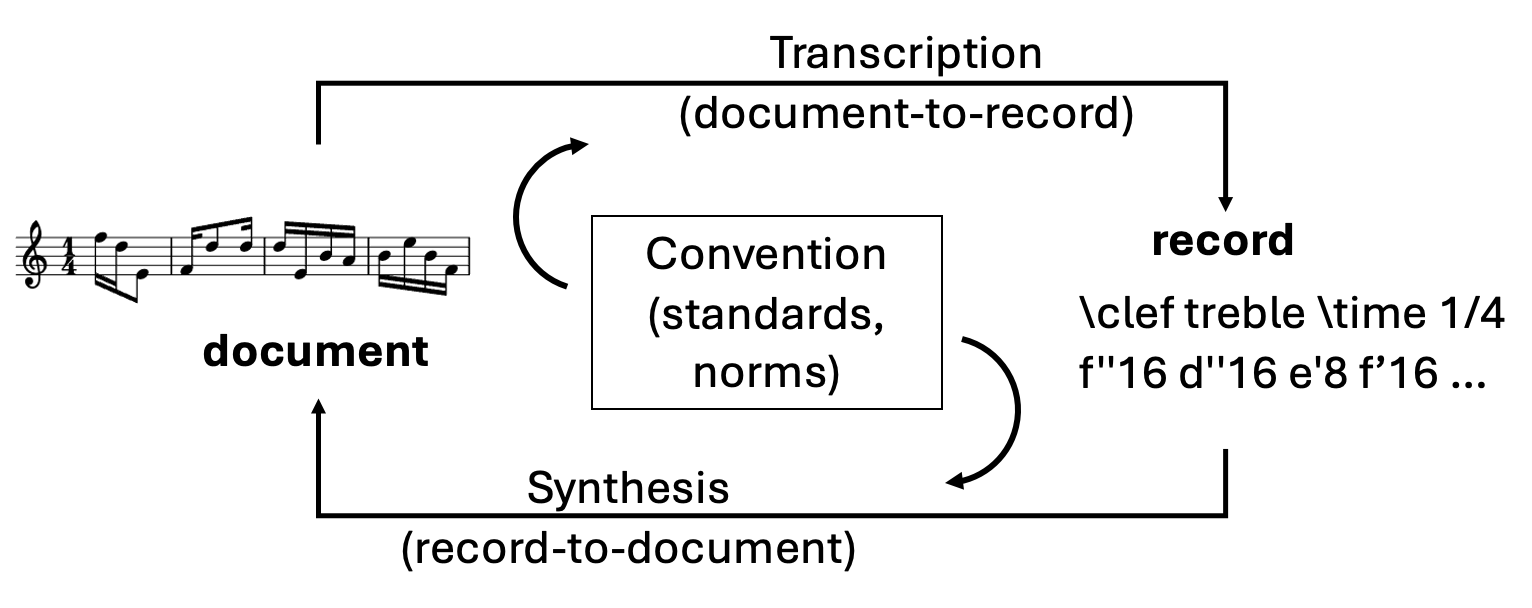}
    \caption{Transcription and synthesis on the example of monophonic sheet music. The record is here represented with LilyPond code \cite{Nienhuys2003LILYPONDAS}.}
    \label{fig:transcription-synthesis}
\end{figure}

% As established in \cref{sec:introduction}, our observed category of convention-bound documents links many common document types conceptually.
% This section formalizes this observation while highlighting that record extraction in convention-bound documents is unambiguous, making the document-to-record transcription a (single-valued) function.

Let $\mathcal{R}$ be the set of records and $\mathcal{D}$ the set of documents for a specific document domain.
A record $r \in \mathcal{R}$ is a property graph, defined as a set of nodes related by a set of edges \cite{angles2018property}.
Specifically, we represent each record as interlinked \emph{(record) nodes}, with each node having a concrete type and a set of type-specific properties.
For example, a music symbol is a record node with a type (e.g., clef or note) and properties (e.g., clef type or note pitch).
Graph theory classifies a (property) graph into certain types based on restrictions present in the graph's relations \cite[Chapter 2.3]{10.5555/22577}.
Based on these types, we define three \emph{record structures} for this work: The sequential record structure for linear graphs and ordered trees (e.g., for sheet music or mathematical expressions), the set record structure for empty graphs (e.g., for shape drawings), and the graph record structure for unrestricted graphs (e.g., for engineering drawings).

To share the information contained in a record $r \in \mathcal{R}$ through a document $d \in \mathcal{D}$, a \emph{write function} $f: \mathcal{R} \rightarrow \mathcal{D}$ first encodes the record into a document (this can also be a human creating the document).
A \emph{read function} $g: \mathcal{D} \rightarrow \mathcal{R}$ decodes the document back to the record (can be a human reading the document).
We call the set of all valid write functions for this document domain its \emph{write function space} $\mathcal{F}\in \mathcal{P}(\mathcal{R} \rightarrow \mathcal{D})$ 
and the set of all valid read functions its \emph{read function space} $\mathcal{G} \in \mathcal{P}(\mathcal{D} \rightarrow \mathcal{R})$, where $\mathcal{P(S)}$ denotes the power set of $S$.

To make the exchange of records via documents precise and efficient, domain-specific conventions exist, such as norms and standards.
These conventions introduce restrictions to (i) what can be expressed (restricting $\mathcal{R}$), (ii) how information must be encoded (restricting the write function space  $\mathcal{F}$), and (iii) how information must be decoded (restricting the read function space  $\mathcal{G}$), such that records can always be exchanged losslessly via documents, meaning the following equation holds:

\begin{equation}
\forall r \in \mathcal{R}, \forall f \in \mathcal{F}, \forall g \in \mathcal{G}: (g \circ f)(r) = r
\label{eq:convention-bound}
\end{equation}

For example, a piece of sheet music can only express what is part of musical notation (restricting $\mathcal{R}$), and standards like the modern staff notation exist to make the exchange lossless for all expressible music pieces (restricting $\mathcal{F}$ and $\mathcal{G}$).

We call documents from domains for which such domain-specific restrictions apply \emph{convention-bound documents}. %, e.g., through standardized symbols.
From \cref{eq:convention-bound} it follows that any convention-bound document $d \in \mathcal{D}$ has only a single, valid reading $r = g^*(d)$ (otherwise \cref{eq:convention-bound} would be violated for the respective records, see \cref{sec:app:single-function} for the full proof).
% or interpretation (otherwise \cref{eq:convention-bound} would be violated). 
Thus, the read function space for $d$ reduces to a single function $\mathcal{G} = \{g^*\}$ in the case of $d$ being a convention-bound document.
This implies that reading convention-bound documents is a many-to-one problem, which permits the use of function approximation techniques that directly estimate the single valid output, rather than having to approximate the (multi-modal) probability distribution over many possible outcomes as in many-to-many problems \cite{bishop1994mixture}.
Importantly, there can still exist many write functions, as any write function $f \in \mathcal{F}$ is valid that encodes the complete record $r$ into the document $d$ in an unambiguously extractable way given $g^*$.

Thus, many valid convention-bound documents  $\mathcal{F}(r) = \{f(r)\ |\ f \in \mathcal{F}\}$ exist for one record, but any of these documents is derived from the same single record, the one causing it.
We hence call the process implemented by any $f \in \mathcal{F}$ \emph{synthesis} and the process implemented by the single read function $g^*$ \emph{transcription} (cp. \cref{fig:transcription-synthesis}).
This terminology is analogous to speech synthesis and transcription: Synthesis indicates the addition of representational variation that does not alter the information content -- there it is a speaker's voice, here it is visual attributes in documents (e.g., stylistic choices like font, line thickness, or semantically neutral symbol placement choices).
Meanwhile, transcription indicates that we ignore these attributes and only extract the core information, the record $r$.

\subsection{Document-to-record transcription} \label{sec:framework}

To put our perspective into practice, we assume the following components in a document recognition as transcription-framework (our concern herein is to machine-learn the transcription model):
A \emph{data engine} provides representations of valid records $r \in R$ as training data for supervised learning, e.g., source code describing the record.
A \emph{synthesis function} $f \in \mathcal{F}$ converts such a representation of the record $r$ into a rasterized document $d = f(r)$, i.e., into a document created for human consumption. When synthesizing training data later, we select synthesis functions strategically from a variety of domain-specific rendering routines to ensure visual diversity, leveraging tools originally developed to visualize records for humans.
A \emph{transcription model} $\hat{g}^*$ converts the document $d$ into a record prediction $\hat{r} \in \mathcal{R}$. 
This transcription model is learned end-to-end (document-to-record), approximating the unknown transcription function $g^*$.
The loss is defined in record space $\mathcal{R}$ (cf. JEPA \cite{lecun2022path,schmidhuber1993discovering}) and measures the dissimilarity between the original record $r$ and its prediction $\hat{r}$.
Measuring the dissimilarity between $r$ and $\hat{r}$ provides a stable learning signal as $r$ is unambiguous for a convention-bound document $d$, meaning any dissimilarity is a mistake of the transcription model (see \cref{sec:formalization}, \cref{eq:convention-bound}).

\begin{figure}[tb]
    \centering
    \includegraphics[width=\linewidth]{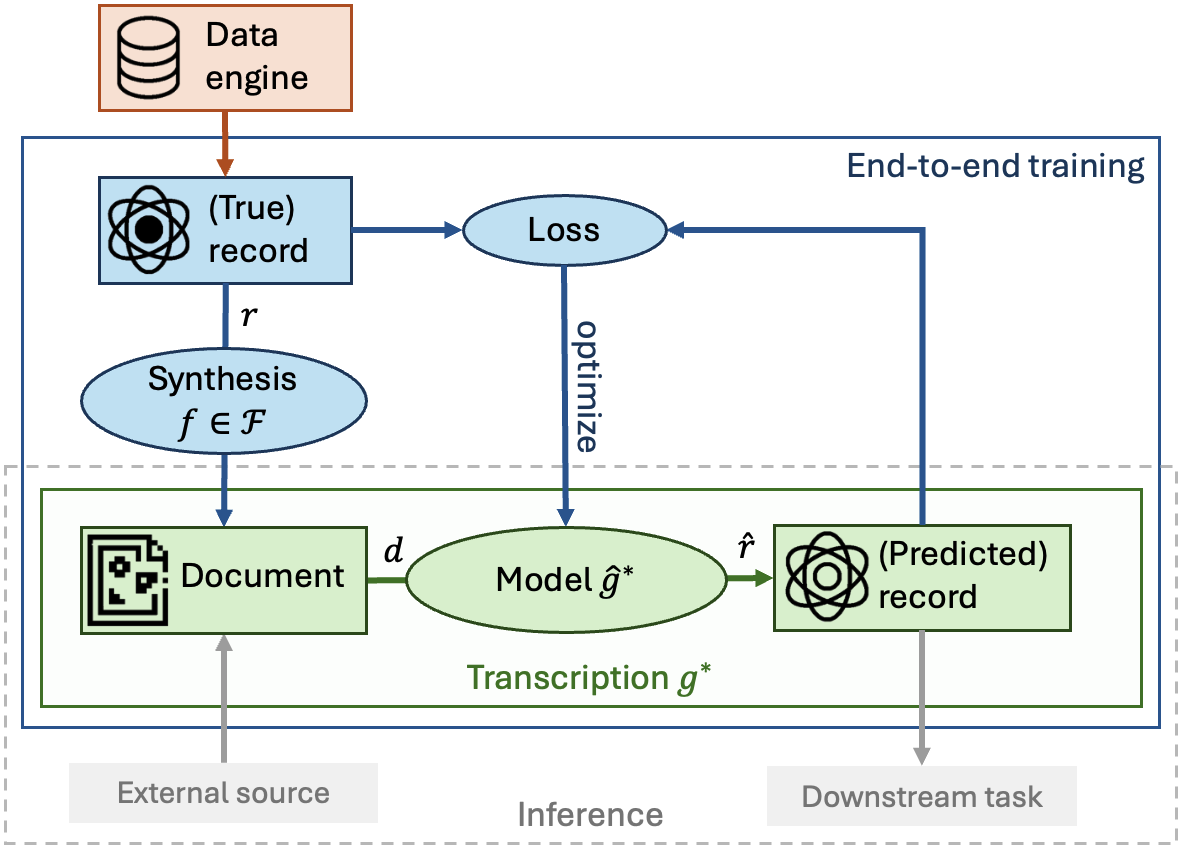}
    \caption{Overview of the proposed document-to-record transcription framework.}
    \label{fig:framework}
\end{figure}

\Cref{fig:framework} illustrates the proposed framework.
For any downstream task, the trained transcription model outputs records that capture all information in a semantically accessible form, omitting all visual by-products of the document and simplifying subsequent processing.
The predicted record can also be synthesized back to a human-friendly document representation via $f(\hat{r})$ for manual transcription validation.

\subsection{Base neural net architecture} \label{sec:base-architecture}

\begin{figure*}[p]
\centering
\begin{subfigure}{\textwidth}
    \includegraphics[width=\textwidth, keepaspectratio]{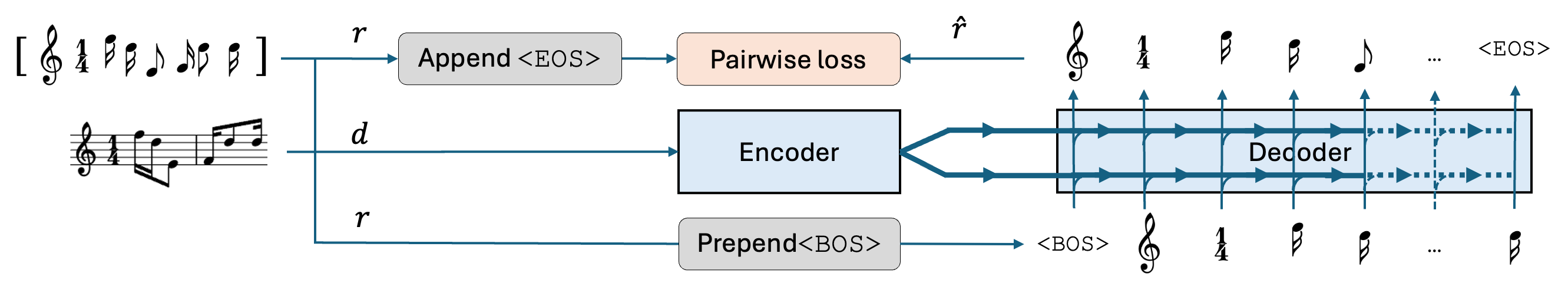}
    \caption{
The training process for next-node prediction in the \emph{document-to-sequence} model (see \cref{sec:seq-to-seq}) implements the standard procedure used in sequence-to-sequence training \cite{sutskever2014sequence} of
teacher forcing \cite{williams1989learning} by prepending a \texttt{<BOS>} token to the decoder input and appending a \texttt{<EOS>} token to the end of the target sequence. Left: Input is the document ($d$); the true record ($r$) serves as ground truth. Right: The decoder predicts, given the current node (bottom), the next node in the sequence (top, thin up-arrows in the decoder) through a pairwise node loss.
Thick arrows in the decoder indicate the information flow through attention masking.
%Pairwise object loss respects order in prediction, meaning the next object must be predicted.
    } \label{fig:seq-to-seq-train}
\end{subfigure}
\begin{subfigure}{\textwidth}
    \includegraphics[width=\textwidth, keepaspectratio]{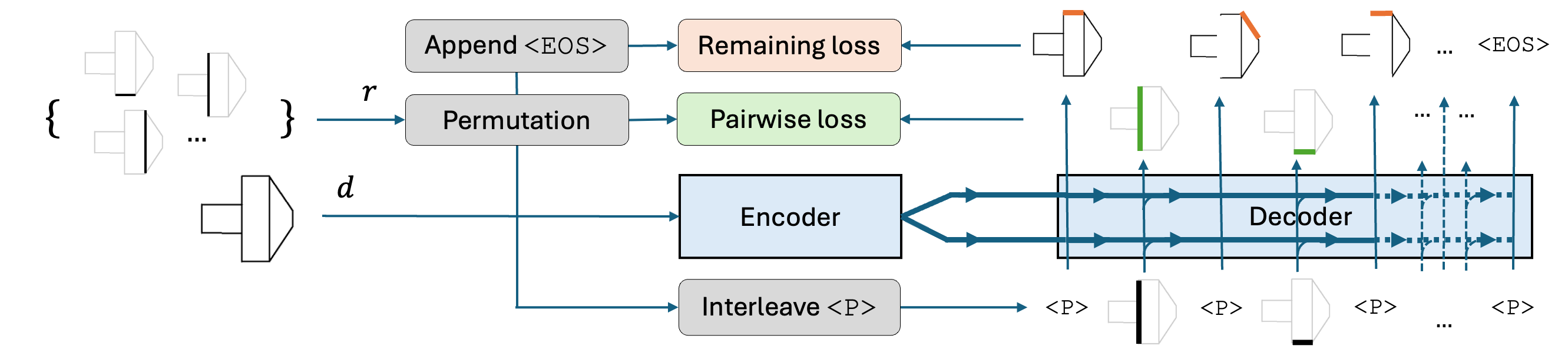}
    \caption{
The training process for remaining-node prediction in the \emph{document-to-set} model (\cref{sec:seq-to-set}) employs teacher forcing by generating a random permutation of the nodes.
To facilitate this, a \texttt{<P>} token is interleaved within the decoder inputs, and a \texttt{<EOS>} token is appended to the targets (see main text for explanation).
Thick arrows in the decoder indicate the information flow through attention masking; no attention on \texttt{<P>} embeddings.
The remaining-node loss respects that there is no order in prediction, meaning any non-``taken'' node can be predicted.
Pairwise loss on ``taken'' nodes encourages information preservation.
} \label{fig:seq-to-set-train}
\end{subfigure}

\begin{subfigure}{\textwidth}
    \includegraphics[width=\textwidth, keepaspectratio]{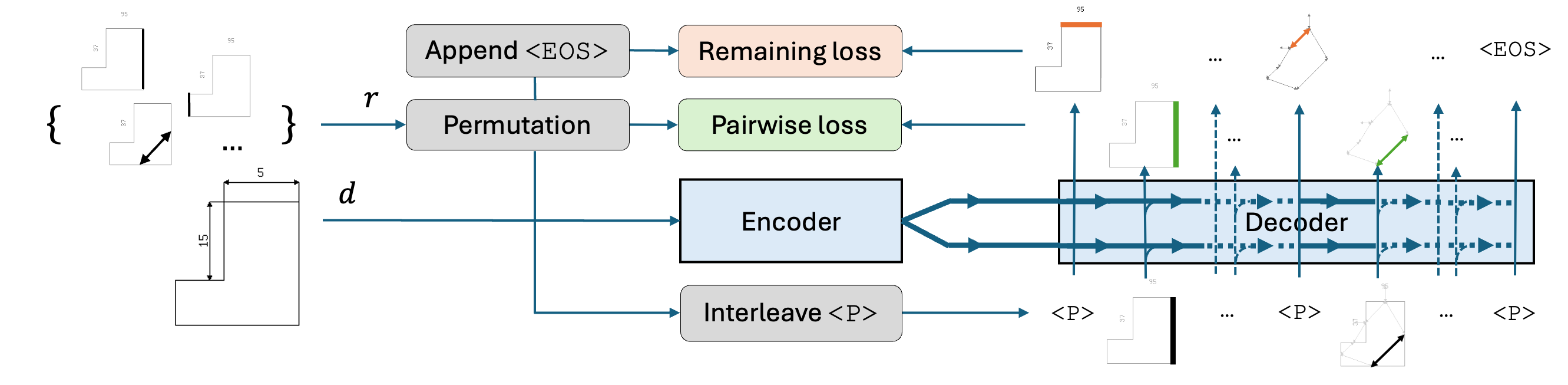}
    \caption{
The remaining-node prediction training process for \emph{document-to-graph} (\cref{sec:seq-to-graph}):
It is equivalent to the \emph{sequence-to-set} training process in \cref{fig:seq-to-set-train} except it additionally guarantees that relationship nodes occur after record nodes in the permutation step.
Relationship nodes are visualized by double arrows that link two record nodes as in \cref{fig:l_shape_example-c}. 
} \label{fig:seq-to-graph-train}
\end{subfigure}
\caption{
Overview of the core components of the architecture and specific adaptations to the training designed for handling distinct record structures.
\texttt{<BOS>}: beginning of record, \texttt{<EOS>}: end of record, \texttt{<P>} prediction token.
Note: For visual clarity, mapping from tokens and nodes to embeddings is not shown.
}
\label{fig:seq-to-record-train}
\end{figure*}

We follow the encoder-decoder paradigm \cite{cho2014learning,sutskever2014sequence,vaswani_attention_2017}, encoding the document $d$ into an intermediate representation to then decode (transcribe) it into the record prediction $\hat{r}$. 
In our experiments, documents contain exactly one page, represented as a single rasterized image.
We make this simplification without loss of generality, as the encoder can be trivially extended to multi-page documents.
The encoder setup is unchanged across all experiments. 
We then add a suitable inductive bias to the decoder and training process based on the record structure as described in the following sections. Each section describes an adaptation in detail (see summary in \cref{fig:seq-to-record-train}).
All adaptations are domain-agnostic by being transferable to any convention-bound document type with the same inherent record structure as the exemplary document type.

\textit{Encoder.} We use a standard vision transformer \cite{dosovitskiy_image_2021} with a minor improvement for digital documents:
Similar to the idea of masked autoencoders, we remove uninformative patches that contain only white pixels \cite{he_masked_2022}.
This reduces the memory footprint and increases computational efficiency without loss in performance.

% \textit{Decoder.} We use the transformer architecture \cite{vaswani_attention_2017} for the decoder, as the base transformer is equivalent to a fully connected graph attention network \cite{velivckovic2017graph} suited to learn a priori unknown relations \cite{velivckovic2023everything}. 
\textit{Decoder.} We use the transformer architecture \cite{vaswani_attention_2017} for the decoder. Disregarding implementation details like layer normalization and positional encoding, the base transformer is equivalent to a fully connected graph attention network \cite{velivckovic2017graph} well-suited for learning \textit{a priori} unknown relations \cite{velivckovic2023everything,lee2019set}.
While we introduce causal attention to enable autoregressive generation, the attention over the preceding context remains unconstrained, ensuring the model captures relations without imposing order-dependent priors.
%, while additionally taking advantage of parallelizable operations on current hardware \cite{vaswani_attention_2017}.
% In the decoder, a record object acts as the base unit, being the analog to a token:
To form the decoder inputs, we embed each record node into a \emph{node embedding} by mapping the record node's type and type-specific properties into a single embedding vector.
Thus, the decoder operates at the level of node embeddings, which is necessary to enable the straightforward inductive bias adaptations discussed in subsequent sections.
% Keeping the unit of an object across the decoder's embeddings makes it easier to add an inductive bias based on the record structure.
After the decoder, output heads predict the type and properties of a record node from each final node embedding.

\subsection{Document-to-sequence bias} \label{sec:seq-to-seq}

For document types with an inherently sequential record structure, the document-to-record task is reduced to the well-known document-to-sequence task.
This section discusses how an inductive bias for sequential structure is built into standard sequence-to-sequence methods, serving as a basis for integrating more complex record structures in later sections.

% \textit{Sequential inductive bias.} 
The seminal transformer paper \cite{vaswani_attention_2017} applied the transformer in a sequence-to-sequence setting (text translation) with the decoder and training process including a sequential inductive bias: \emph{next-token prediction}.
We do the equivalent \emph{next-node prediction} for our sequential record structure.
% each object embedding 
Specifically, at each position, the model predicts the next node in the record sequence based on the current and previous node embeddings.

\begin{figure*}[tb]
\centering
\begin{subfigure}{0.9\textwidth}
    \includegraphics[width=\textwidth, keepaspectratio]{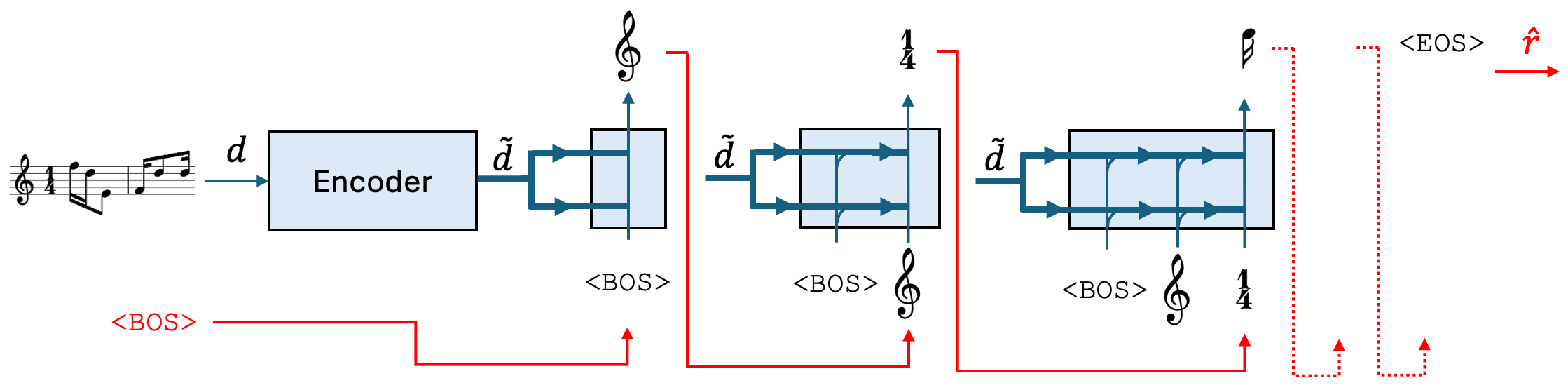}
    \caption{Sequence-to-sequence inference process: Predict one \emph{next} node at a time.}
    \label{fig:seq-to-seq-infer}
\end{subfigure}
\hfill
\begin{subfigure}{0.9\textwidth}
    \includegraphics[width=\textwidth, keepaspectratio]{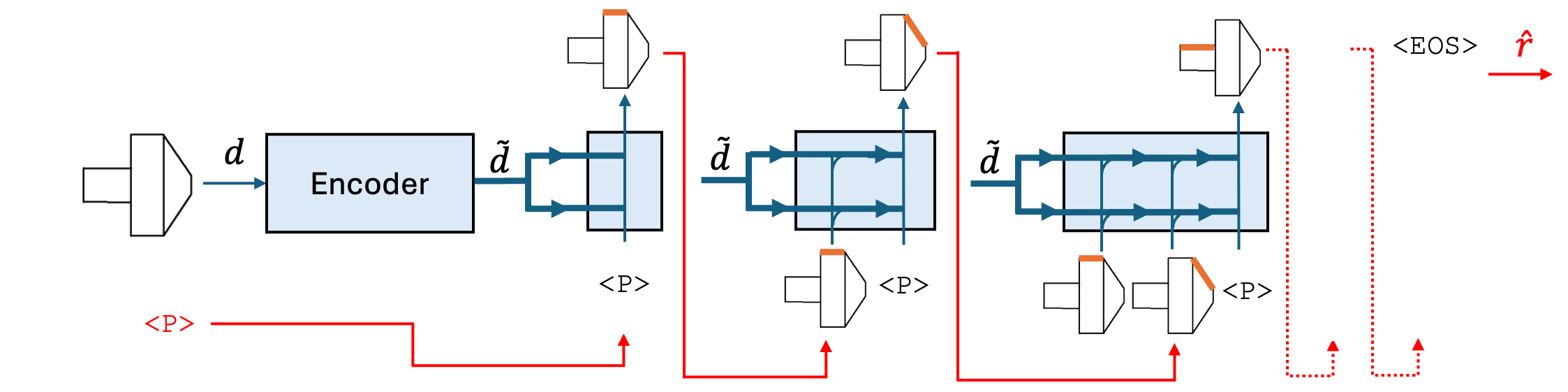}
    \caption{Sequence-to-set inference process: Predict one \emph{remaining} node at a time.}
    \label{fig:seq-to-set-infer}
\end{subfigure}
\caption{
The proposed sequence-to-set inference process (bottom) compared to the well-known sequence-to-sequence inference process (top).
Inference is basically the same autoregressive mechanism, where sequence-to-sequence prepends the \texttt{<BOS>} token, while sequence-to-set appends the \texttt{<P>} token.
% Conceptually, the sequence is predicted in order, one \emph{next} object at the time, whereas the set is predicted one \emph{remaining} object at the time.
} \label{fig:infer}
\end{figure*}

We do this following the common approach: 
The decoder input is the node sequence prepended with a special \emph{beginning-of-record} token \texttt{<BOS>}, and the decoder output target is the node sequence appended with a special \emph{end-of-record} token \texttt{<EOS>}.
The loss is a node dissimilarity measure (node type and properties) at each position (see \cref{eq:seq-seq-loss}).
Causal masking, meaning embeddings can never attend to the right, is applied throughout the decoder, preventing trivial label leakage solutions.
One-dimensional positional encoding is added to the initial node embeddings to enable the decoder to learn and use the sequential structure of the nodes.
\Cref{fig:seq-to-seq-train} illustrates this training process.
Inference happens autoregressively, predicting one next node at a time (see
\cref{fig:seq-to-seq-infer}).

Formally, the sequence bias loss is defined as follows:
Let $r$ be a sequence of $m$ nodes $(n_1, ..., n_m)$. Let $n_0$ be the \texttt{<BOS>} token and $n_{m+1}$ be the \texttt{<EOS>} token (which, for notational convenience, is treated as a property-less node with unique type). Let $\hat{n}_{i} = \hat{g}^*(d,n_0,\hdots,n_{i-1})$ be the prediction of the next node $n_{i}$ based on the document $d$, the \texttt{<BOS>} token $n_0$ and previous nodes $(n_1,\hdots, n_{i-1})$; $\hat{g}^*$ representing our model. Let $l(n,\hat{n})$ be a dissimilarity measure between a node $n$ and a prediction $\hat{n}$ based on their type and properties (see \cref{sec:node-dissimilarity} for definition). Then, the sequence bias loss $\mathcal{L}_{seq}$ is defined as:
\begin{equation}
    \mathcal{L}_{seq} = \sum_{i=1}^{m+1} l(n_{i}, \hat{n}_{i})
    \label{eq:seq-seq-loss}
\end{equation}

\subsection{Document-to-set bias} \label{sec:seq-to-set}

%By assuming a sequential record structure, the document-to-record task reduces to a document-to-sequence task.
%We integrate this sequential record structure as an inductive bias into the decoder and training process.
%We use simplified sheet music documents as an exemplary document type.

For document types with a set record structure, the document-to-record task becomes a document-to-set task.
This section explains how to adapt the document-to-sequence model to the document-to-set task by replacing the sequence bias of next-node prediction with a set bias, termed here \emph{``remaining-node prediction''}.
We later apply this method to shape drawings as an exemplary document type.

% \textit{Set inductive bias.}
For a set structure, the sequential bias of next-node prediction makes no sense as there is no natural notion of a ``next node'' in an unordered set of nodes.
Therefore, we replace next-node prediction with remaining-node prediction: This means that from each final node embedding (after attending to the encoded document and previous node embeddings within the decoder), we predict \emph{any} record node visible in the document but not ``taken'' by the previous node embeddings.
Previous nodes are defined by the decoder's input, which is a random permutation of the record nodes during training and autoregressive predictions during inference.
This is the main idea of the adaptation to a set record structure; however, for it to successfully work in a transformer-based decoder, a decisive technical detail is needed.

A critical part of the success behind transformer-based next-token prediction is to significantly speed up training by predicting the next token at each position simultaneously in a single forward pass, based on the previous ground-truth tokens, a technique known as teacher forcing \cite{williams1989learning}.
However, this becomes problematic for (naive) remaining-node prediction:
Due to this parallel prediction, each node embedding is transformed, layer by layer, to become its required prediction (here, one of its remaining nodes), while, at the same time, previous node embeddings are transformed to become their required predictions (here, one of their remaining nodes).
Yet, if the previous node embeddings are transformed, the original ``taken'' node information is lost. 
Thus, it becomes impossible for later node embeddings to know which nodes are remaining, that is, which nodes they can predict, and training fails.
This issue highlights that embeddings have two roles in a transformer-based decoder: (a)~they must be transformed to become their required prediction and (b)~they must carry the (contextualized) original information for the other embeddings. He et al. \cite{he2025law} empirically validate this conflict, showing through probing that token embeddings gradually transition from representing input information to representing the prediction target as they move through the layers.
% This seems to be not relevant for next-token prediction, as the original information doesn't get lost as the previous embedding must predict it.

For remaining-node prediction to work successfully, we must separate these two roles into two separate embeddings: 
one prediction embedding that, in our case, becomes a remaining node (role  ``a'') and another node embedding that carries the ``taken'' node information for the other embeddings (role ``b'').
This happens to be similar to the architectural adaptations of XLNet \cite{yang2019xlnet}; however, we do it for (remaining-node) set prediction, whereas XLNet is motivated to learn (autoregressive, bidirectional) sequence prediction (cf. \cite{pannatier2024sigma}).
Specifically, we add one prediction token \texttt{<P>} in front of each node, allowing for parallel prediction at the cost of doubling the decoder's sequence length, and add one extra prediction token \texttt{<P>} at the end to predict a special end-of-record token \texttt{<EOS>} as a stop signal during autoregressive inference. 
Each prediction token is then mapped to a prediction embedding, which, with the node embeddings, forms the decoder's input.
Inside the decoder, prediction and node embeddings can only attend to previous or ``taken'' node embeddings; prediction embeddings are never attended to as they carry information irrelevant to other embeddings.
From each final prediction embedding, we predict a remaining node by being subject to a \emph{remaining-node loss} that is the minimum dissimilarity over all remaining nodes (see \cref{eq:set-loss-rem}).
From each final node embedding, we predict its original input node, encouraging the model to preserve (or pass through) the ``taken'' node information necessary for other embeddings (see \cref{eq:set-loss-taken}).
\Cref{fig:seq-to-set-train} illustrates this training process.
For the set inductive bias, we do not add positional encoding, as there is no sequential order in the node embeddings.
Inference happens autoregressively, predicting one remaining node at a time (see \cref{fig:seq-to-set-infer}).

Formally, the set bias loss is defined as follows:
Let $r$ be a permutation of a set of $m$ nodes $(n_1, ..., n_m)$. Let $p$ be the prediction token \texttt{<P>} and $n_{m+1}$ be the \texttt{<EOS>} token (which, for notational convenience, is treated as a property-less node with unique type). Let $\hat{n}_{i} = \hat{g}^*(d,p,n_1,\hdots,n_{i-1})$ be the prediction of a remaining node based on the document $d$, the prediction token $p$ and the taken nodes $(n_1,\hdots,n_{i-1})$. Let $\tilde{n}_{i} = \tilde{g}^*(d, n_1,\hdots,n_{i-1},n_i)$ be the prediction (or pass-through) of a taken node $n_{i}$ based on the document $d$, the actual node $n_i$ and previous taken nodes $(n_1,\hdots,n_{i-1})$; $\hat{g}^*$ and $\tilde{g}^*$ representing our model. Let $l(n,\hat{n})$ be the same dissimilarity measure between a node $n$ and a prediction $\hat{n}$ as above. The loss $\mathcal{L}_{set}$ is defined as the loss for the remaining-node predictions $\mathcal{L}_{rem}$, the loss to predict the \texttt{<EOS>} token correctly $\mathcal{L}_{eos}$, and the loss for passing through the taken nodes $\mathcal{L}_{taken}$:
\begin{subequations}
\begin{gather}
\mathcal{L}_{set} = \mathcal{L}_{rem} + \mathcal{L}_{eos} + \mathcal{L}_{taken} \\
\mathcal{L}_{rem} = \sum_{i=1}^m \min_{j=i}^m l(n_{j}, \hat{n}_{i}) \label{eq:set-loss-rem} \\
\mathcal{L}_{eos} = l(n_{m+1},\hat{n}_{m+1}) \\
\mathcal{L}_{taken} =  \sum_{i=1}^m l(n_{i}, \tilde{n}_{i}) \label{eq:set-loss-taken}
\end{gather}
\end{subequations}

\subsection{Document-to-graph bias} \label{sec:seq-to-graph}

For document types with records best captured in an unrestricted graph structure, the document-to-record task becomes a document-to-graph task.
In this section, we adapt the document-to-set model of the previous section to the document-to-graph task by forming an implicit graph from a set structure containing nodes for both record nodes and their relationships (edges).
We assume no order in the nodes or edges, but differentiate between them by first predicting all record nodes.
We later apply this method to simplified engineering drawings as an exemplary document type.

% \textit{Graph inductive bias.} 
We represent the record graph as an implicit graph induced by a relational model \cite{10.1145/362384.362685,boudaoud2022towards}: Each record node gets a unique identifier and relations between nodes are represented as additional \emph{``relationship nodes''} with properties to link record nodes by their identifier.
For example, an undirected graph of three nodes $\{A, B, C\}$ and an edge from $A$ to $B$ will be represented as a set of four nodes including the additional relationship node $Z$ with two nodes as properties: $\{A, B, C, Z(A, B)\}$.
This represents a graph as a set of nodes indirectly related through their properties, allowing the use of the same architecture as in \cref{sec:seq-to-set} for the processing of any graph.

During training, when generating permutations of the record, we ensure that all record nodes appear before all relationship nodes, guaranteeing that the record nodes and their identifiers are known when predicting their relationships (see \cref{fig:seq-to-graph-train}).
We use the position in the decoder sequence as the identifier of a node and add one-dimensional positional encoding to provide this positional information to the model with a prediction embedding sharing the same positional encoding as the node embedding to its right.
Given the above graph example, we generate, e.g., the following permutations during training: $[A, B, C, Z(0, 1)]$, $[C, B, A, Z(1, 2)]$.
This setup allows the model to first predict the set of record nodes visible in the document (symbol recognition) and then predict their specific relationships (symbol assembly).

Formally, the graph bias loss is defined as follows:
Let $r$ be two concatenated sequences: A sequence of $m$ record nodes $(n_1, ..., n_m)$ and a sequence of $o$ relationship nodes $(n_{m+1}, ..., n_{m+o})$. Let $p$ be the prediction token \texttt{<P>} and $n_{m+o+1}$ be the \texttt{<EOS>} token (which, for notational convenience, is treated as a property-less node with unique type). Let $\hat{n}_{i} = \hat{g}^*(d,p,n_1,\hdots,n_{i-1})$ be the prediction of a remaining node based on the document $d$, the prediction token $p$ and the taken nodes $(n_1,\hdots,n_{i-1})$. Let $\tilde{n}_{i} = \tilde{g}^*(d,n_1,\hdots,n_{i-1},n_i)$ be the prediction (or pass-through) of the taken node $n_{i}$ based on the document $d$, the actual node $n_i$ and previous taken nodes $(n_1,\hdots,n_{i-1})$; $\hat{g}^*$ and $\tilde{g}^*$ representing our model. Let $l(n,\hat{n})$ be the same dissimilarity measure between a node $n$ and a prediction $\hat{n}$ as above. The loss $\mathcal{L}_{graph}$ is defined as the loss for the remaining-record-node predictions $\mathcal{L}_{rem}^{nodes}$, the remaining-relationship-node predictions $\mathcal{L}_{rem}^{rel}$, the loss to predict the \texttt{<EOS>} token correctly $\mathcal{L}_{eos}$, and the loss for passing through the taken nodes $\mathcal{L}_{taken}$:
\begin{subequations}
\begin{gather}
\mathcal{L}_{graph} = \mathcal{L}_{rem}^{nodes} + \mathcal{L}_{rem}^{rel} + \mathcal{L}_{eos} + \mathcal{L}_{taken} \\
\mathcal{L}_{rem}^{nodes} = \sum_{i=1}^m \min_{j=i}^m l(n_{j}, \hat{n}_{i}) \\
\mathcal{L}_{rem}^{rel} = \sum_{i=m+1}^{m+o} \min_{j=i}^{m+o} l(n_{j}, \hat{n}_{i}) \\
\mathcal{L}_{eos} = l(n_{m+o+1},\hat{n}_{m+o+1}) \\
\mathcal{L}_{taken} =  \sum_{i=1}^{m+o} l(n_{i}, \tilde{n}_{i})
\end{gather}
\end{subequations}

\section{Experiments}

\subsection{Overview}

To show the validity of our approach, we conduct experiments on three archetypal document types for the progression of simple to more advanced record structures discussed above (i.e., sequence, set, and graph):  We introduce how we represent records and create respective training data for monophonic sheet music (\cref{sec:sheet-music-setup}), shape drawings (\cref{sec:shape-drawings-setup}), and simplified engineering drawings (\cref{sec:engineering-drawings-setup}).
We then detail the common experimental setup for model training and evaluation in \cref{sec:training-setup}.
We report results in \cref{sec:results-appropriate} and provide an ablation study that sheds light on the influence of an appropriate record structure bias in \cref{sec:results-inappropriate}.

\subsection{Sequence bias: Sheet music} \label{sec:sheet-music-setup}

We use monophonic sheet music documents as an exemplary document type for the document-to-sequence task.

\textit{Record representation.} The sheet music record is a sequence of one clef node, one time signature node, followed by music note nodes. The clef node and the time signature node have one discrete property: the clef type and the time signature. Each music note node has two discrete properties: duration and pitch (vertical note position).

\textit{Data engine.} For training, we generate \num{40000} synthetic, simplified (monophonic, four-bar, single-staff) music sheets.
We generate two possible clef types (treble \musSymbolClef{\symbol{72}} or bass \musSymbolClef{\symbol{74}}), four possible time signatures (1/4 \musMeter{1}{4}, 2/4 \musMeter{2}{4}, 
3/4 \musMeter{3}{4}, and common time \meterC), five possible note durations (whole \musWhole, half \musHalf, quarter \musQuarter, eighth \musEighth, and sixteenth \musSixteenth) and nine possible pitches spanning five staff lines and four in-between staff spaces. All property values are chosen uniformly at random, with note duration respecting the remaining duration of the current bar.

\textit{Synthesis.} We render each synthetically generated staff onto a raster image. The image is $100$ pixels high and has a flexible width to guarantee that all music notes are rendered on one single staff line. \Cref{fig:music_data_example} shows an example.

\begin{figure}[hb]
\centering
    \includesvg[width=\linewidth]{img/music_example.svg}
    \caption{Example of a simple sheet music document used for training. \Cref{fig:music_data_examples} shows more examples.}
    \label{fig:music_data_example}
    \vspace{-1cm}
\end{figure}

\subsection{Set bias: Shape drawings} \label{sec:shape-drawings-setup}

We use shape drawings as an exemplary document type for the document-to-set task.

%\smallskip \noindent
\textit{Record representation.} We represent shape drawings as a set of record nodes representing simple geometric shapes. Without loss of generality, our simplified proof-of-concept implementation will use a line node and a circle node; others could be added.
Each line node contains two properties, representing the coordinates of the start point and the end point in a tuple. 
Each circle node contains two similar properties as well: the circle's leftmost point and rightmost point (i.e., the extreme points on the horizontal middle axis, implicitly defining also the circle's height).
Each coordinate tuple consists of continuous values of normalized \(x\)- and \(y\) pixel coordinates.

%\smallskip \noindent
\textit{Data engine.} The ABC dataset \cite{koch_abc_2019} contains one million 3D parts as 3D scene graphs.
% a public implementation\footnote{\url{https://dev.opencascade.org/doc/overview/html/index.html}} of
We randomly project \num{300000} of those to 2D edges as seen from one of the six principal views using the hidden line removal algorithm \cite{galimberti_algorithm_1969}.
We clean the projection by removing superimposed edges and rejoining 2D edges that were originally joined in 3D. 
We only take projections containing up to \num{10} circles or lines and no other edges like arcs or b-splines. The final dataset consists of about \num{900000} projections.

\begin{figure}[hb]
    \centering
    \includegraphics[width=\linewidth]{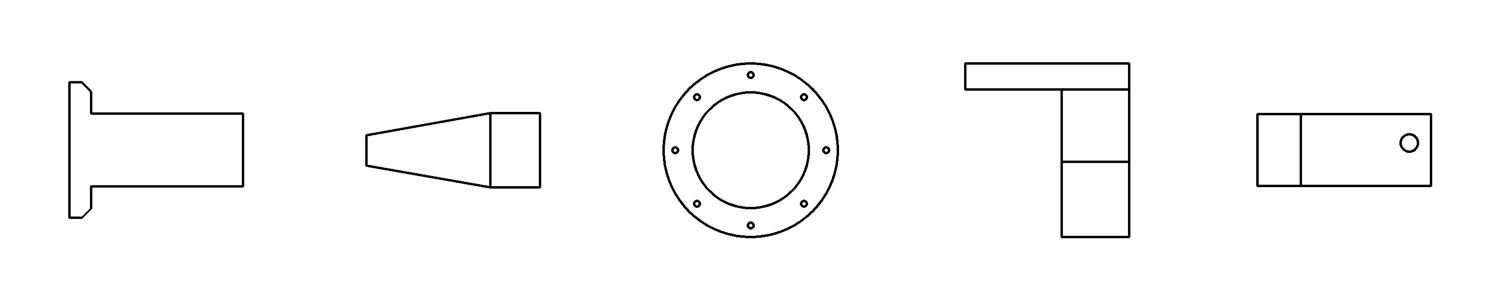}
    \caption{Five examples of simple technical documents, i.e, rendered projections of engineering parts, used for training. \Cref{fig:abc_data_examples} shows more examples.}
    \label{fig:abc_data_example}
\end{figure}

%\smallskip \noindent
\textit{Synthesis.} We render all lines and circles onto a $280 \times 280$ pixel image.
During training, we augment each record with random margins, random translations, random scaling, and axis mirroring while ensuring all lines and circles remain fully visible (see \Cref{fig:abc_data_example}).

\subsection{Graph bias: Engineering drawings} \label{sec:engineering-drawings-setup}

We use simplified engineering drawings as an exemplary document type for the document-to-graph task.

%\smallskip \noindent
\textit{Record representation.} We represent simplified engineering drawings as a record of connected line nodes and dimension annotation nodes (numbers indicating relative line lengths).
Each line node contains two properties, representing the coordinates of the start point and the end point in a tuple. 
Each dimension annotation contains a single property, representing the coordinates of its center in a tuple (stored in the same format as described above).
Relationship nodes are the connections between two lines or the link between a dimension annotation and its line.
% A line connection is a relationship between two lines.
% Each image coordinate tuple consists of continuous values of normalized \(x\)- and \(y\) pixel coordinates.

\begin{figure}[b!]
    \centering
    \begin{subfigure}{0.3\linewidth} % Specify width of subfigure
        \centering
        \includegraphics[width=\textwidth]{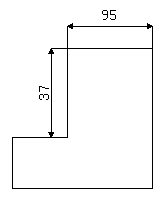}
        \captionsetup{labelformat=empty}
        \caption{(a) Document}
        \label{fig:l_shape_example-a}
    \end{subfigure}\hfill % Add horizontal space between subfigures
    \begin{subfigure}{0.3\linewidth} % Specify width of subfigure
        \centering
        \includegraphics[width=\textwidth]{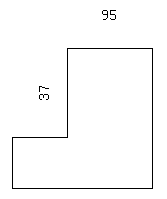}
        \captionsetup{labelformat=empty}
        \caption{(b) Nodes} % Add caption
        \label{fig:l_shape_example-b}
    \end{subfigure}\hfill % Add horizontal space between subfigures
    \begin{subfigure}{0.3\linewidth} % Specify width of subfigure
        \centering
        \includegraphics[width=\textwidth]{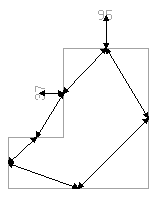}
        \captionsetup{labelformat=empty}
        \caption{(c) Edges} % Add caption
        \label{fig:l_shape_example-c}
    \end{subfigure}
    \caption{
Example of a synthetic engineering drawing. 
Left: The document, where helplines and arrows help to communicate links between dimension annotations and lines. % relations.
Middle, right: A visualization of the pure record, namely the record nodes (middle) and the relationships (edges; see left) by means of arrows.
}
    \label{fig:l_shape_example}
\end{figure}

%\smallskip \noindent
\textit{Data engine.} We create synthetic L-shaped drawings with random augmentations, specifically, mirroring, rotations with multiples of $90$ degrees, and translations.
Each L-shape has six lines, each connected to its neighbors and each having a \SI{30}{\percent} chance of having a dimension annotation. 
\cref{fig:l_shape_example-b,fig:l_shape_example-c} visualize an example.

%\smallskip \noindent
\textit{Synthesis.} We render each L-shape onto a $140 \times 140$ pixel image.
Dimension annotations are rendered as a random number between $0$ and $100$, accompanied by visually guiding helplines and arrows.
\cref{fig:l_shape_example-a} shows an example.
The rendering simulates multiple synthesis functions through randomized line thicknesses, grayscale variations, $3$ arrow styles, $4$ fonts, $10$ font sizes, and random blur augmentation.
\Cref{fig:l_shape_examples} shows examples.

\subsection{General experimental setup} \label{sec:training-setup}

All experiments and models follow the same general setup, and all models have around \num{65} million learnable parameters.

\textit{Encoder:} 
The encoder is a basic vision transformer.
We divide the input image into patches of size $10 \times 10$ pixels, add 2D positional encodings \cite{parmar_image_2018}, remove all uninformative patches (i.e., background-only), and linearly map the flattened informative patches to form the encoder's input. % into the encoder's embedding space.
The encoder is a vision transformer with $3$ layers, a $512$-dimensional embedding space, and $8$-head residual self-attention (unmasked) followed by a residual feed-forward network.
The feed-forward network consists of, respectively, a $2048$-dimensional hidden layer with no bias term and the GELU activation function \cite{hendrycks2016gaussian}, a layer norm layer, a $512$-dimensional output layer with no bias term and linear activation function, and another layer norm layer.
 
\textit{Decoder:} 
The decoder follows the same setup as the encoder, except that it uses cross-attention on the image embeddings from the encoder and masked self-attention based on the above bias adaptations.
\textit{Node encoder:} 
The beginning-of-record token \texttt{<BOS>} and prediction token \texttt{<P>} are each mapped to $512$-dimensional embedding vectors.
A record node is mapped to a node embedding in the following way:
The node type and discrete properties are each mapped to a $64$-dimensional vector. Each coordinate property value is linearly mapped to a $64$-dimensional vector.
The node type and properties are concatenated and linearly mapped into the decoder's embedding space.
\textit{Node prediction head:} 
From a final decoder embedding, we predict the node type and properties. 
The node type and each discrete property are linear maps followed by a softmax activation function. 
Coordinate properties use a novel coarse-to-fine method for pixel-precise prediction: First, we predict the image patch for each coordinate property by calculating a cross-attention probability from the node embedding to the image patches.
Then, we predict the pixel inside the $10 \times 10$ patch by concatenating the most probable patch embedding to the node embedding and linearly map it to \num{100} softmax outputs.

\textit{Loss functions:}
Two loss functions are used as outlined above -- pairwise loss and remaining-node loss.
%The pairwise loss sums up the difference between each prediction and its target node.
%The remaining-node loss sums up the minimum difference of each prediction with any non-``taken'' nodes.
%The difference between prediction and target node is the sum of the cross-entropy for the node type and each property.
All loss concerning the prediction of node properties contribute to the overall loss only if the respective node type was predicted correctly; pixel prediction loss only if the respective patch was predicted correctly.

\textit{Training:}
All models are trained to convergence using their respective domain-specific data engine and a batch size of $32$.
This results in training the sheet music model for \num{31250} steps: \num{25} epochs of \num{40000} training samples; 
the shape drawing model for \num{506250} steps: \num{18} epochs of \num{900000} augmented training samples;
the engineering drawing model for \num{62500} steps: \num{40} epochs of \num{50000} generated training samples.
For optimization, we use the AdamW optimizer \cite{loshchilov_decoupled_2018} with a learning rate of \numprint{1e-4}, $\beta_1 = $ \num{0.9}, $\beta_2 = $ \num{0.98}, $\epsilon = $ \numprint{1e-9}, a clip norm of \num{0.1}, and a weight decay of \num{0.004} based on preliminary experiments. 
To speed up training, we use batch processing and a hardware-friendly structure of array data representation (see \cref{sec:app:training-setup}).

\textit{Evaluation:}
All models are evaluated using set-aside validation data created by the same domain-specific data engine.
Model performance is evaluated using the \emph{transcription accuracy}, defined as the fraction of predictions ($\hat{r}$) that are equal ($\hat{r} = r$) to their target records ($r$).
Records are equal if their nodes, edges, and structure are equal (i.e., a property graph isomorphism, see \cref{sec:record-equality} for a formal definition).
For sequences, this means all nodes in the sequence must be pairwise equal.
For graphs and sets, this means they must contain the same record nodes, the same relationship nodes, and the same structure.
Two nodes are equal if their type is equal and their properties are equal (up to a predefined precision threshold for continuous coordinates).
Thus, transcription accuracy renders the entire transcription incorrect for any partial error, e.g., predicting one property wrong. We believe this to be appropriate as any possible error may have detrimental consequences in some downstream task. Note that this conservative metric renders the task particularly hard, similar to the exact match metric in multi-label classification \cite{ganda2018survey}, which considers a multi-label classification incorrect if any label is wrong or missing.

\subsection{Results} \label{sec:results-appropriate}

In the main set of experiments, we evaluate the performance when the appropriate inherent bias for each given record structure is used.

\textit{Sheet music recognition as document-to-sequence transcription:} \label{sec:res:seq-to-seq}
We use \num{10000} randomly generated single-staff music sheets to evaluate the model adapted to the sequential bias (\cref{sec:seq-to-seq}). 
The model transcribes \num{9659} sheets of music exactly and has a transcription accuracy of \SI{96.6}{\percent}. This is hard to compare exactly to the more forgiving metrics like IoU usually used in OMR \cite{tuggener_real_2024}, but appears on par with the state of the art for this relatively simple task.

\textit{Shape drawing recognition as document-to-set transcription:}
For evaluation, we generate \num{10000} 2D projections of separate 3D parts from the ABC data set.
Our model transcribes \num{7490} projections correctly up to a coordinate precision threshold of about $\pm4$ pixels. This results in a transcription accuracy of \SI{74.9}{\percent}, indicating that our method successfully learned the document-to-set task end to end.

\textit{Engineering drawing recognition as document-to-graph transcription:}
We use \num{1000} randomly generated L-shapes using the same synthetic data engine as for training. 
The model (with graph bias) transcribes \num{748} L-shapes correctly up to a coordinate precision threshold of about $\pm4$ pixels. This results in a transcription accuracy of \SI{74.8}{\percent}.
% To the authors' knowledge, this is the first time the transcription of an inherently non-sequential document type has ever been successfully learned end to end.
To the authors' knowledge, this is the first time that the transcription of mechanical engineering drawings has been successfully learned end-to-end, and only recently has successful end-to-end learning of any inherently non-sequential document type been reported for the first time \cite{sturmer2025engineering,hu2024raster}. 

The above results are proof of principle that our method of inductive bias adaptation works, i.e., it enables a model to successfully learn the document-to-sequence, document-to-set, and document-to-graph tasks.

\subsection{Ablation studies: Choosing inappropriate biases} \label{sec:results-inappropriate}
In this section, we conduct an ablation study to determine whether the transcription performance above is actually due to our method of bias adaptation by deliberately using a non-optimal bias for certain cases: The set and graph inductive bias in case of an inherently sequential record structure and the sequential inductive bias in case of an inherently set or graph record structure. This last setup basically replicates what has implicitly been done in much of past document recognition works.

\textit{Treating sheet music with a set bias:}
The model architecture built for the set inductive bias (\cref{sec:seq-to-set}) is adapted to the sheet music scenario (\cref{sec:seq-to-seq}) with its inherently sequential record structure.
As this architecture predicts music symbols in no particular order, we must add a horizontal position property to each record node, representing its position in the sequence, to recover the actual sequential record at inference time. This approach effectively encodes a sequence as a set, as suggested in \cite{vinyals2015order}.
Using the set inductive bias leads to a more difficult task, as the model must learn the sequential nature on its own without structural guidance through an appropriately biased model architecture. For example, the prediction of the last record node requires checking all ``taken'' nodes against all symbols visualized in the document. 
The model transcribes \num{1937} sheets of music exactly and thus has a transcription accuracy of \SI{19.4}{\percent}, dropping \num{77.2} percentage points compared to the model with sequential bias.
The model performs especially badly on longer sequences with a transcription accuracy of \SI{0.0}{\percent} for more than $15$ symbols.
Similarly, the model architecture employing the general graph inductive bias (\cref{sec:seq-to-graph}) yields \SI{0.0}{\percent} accuracy under the baseline experimental setup.
The set and graph bias make fewer assumptions about the underlying graph structure than the sequential bias. While this makes the training less data-efficient, more data would enable the model (or a bigger model) to eventually learn the inherent structure and solve such transcription tasks.
For example, extending the experimental training setup for the same graph model to \num{500} epochs (and increasing batch size to \num{512}) improves the transcription accuracy to \SI{10.4}{\percent}.

% \textit{Treating sheet music with a graph bias:}
% The model architecture build for the general graph inductive bias (\cref{sec:seq-to-graph}) is applied to the sheet music scenario (\cref{sec:seq-to-seq}) with its inherently sequential record structure.
% Using the general graph bias leads to a more difficult task, as the model must learn to predict the set of music notes followed by the relations rather than predicting a sequence of notes directly.
% With the same training setup of \num{25} epochs with a batch size of \num{32}, the model transcribes \num{0} sheets of music exactly, as it has not converged. However, training the model for \num{20} epochs leads to convergence and allows the model to predict \num{15} sheets of music exactly. Additionally, increasing the batch size to $512$ and training it for an additional \num{200} epochs, lets it transcribe \num{000} sheets of music exactly and thus has a transcription accuracy of \SI{000}{\percent}. 
% Conceptually, that the initial results trail those of the model with sequential bias is expected, as the sequential order must be learned. Empirically, however, these preliminary results show that scaling our graph inductive bias to more complex scenarios is possible.

\textit{Treating shape drawings with a sequence bias:}
The model architecture built for the sequential inductive bias (\cref{sec:seq-to-seq}) is adapted to the shape drawing scenario (\cref{sec:seq-to-set}) with its inherent set record structure.
Using the sequential bias leads to a more difficult task, as the model must learn the uniform distribution over the remaining nodes rather than a single node, hardened by record nodes carrying a type and properties.
The model transcribes \num{1193} drawings correctly up to a coordinate precision threshold of about $\pm4$ pixels, resulting in a transcription accuracy of \SI{12.0}{\percent}, dropping \num{62.9} percentage points compared to the model with set bias.
The model performs especially badly on records with more than four record nodes, making only \num{23} correct transcriptions and a transcription accuracy of \SI{0.6}{\percent} on such records.
The sequential bias makes more (here: wrong) assumptions about the underlying graph structure. These assumptions lead to a noisy training signal, as the predicted next node can be any remaining node. We suspect that this noisy signal leads to unstable training, making more data insufficient for learning non-sequential transcription tasks. 
Similarly, adapting this misaligned bias to the engineering drawings scenario (\cref{sec:seq-to-graph}), which possesses a general graph structure, fails entirely, yielding \num{0} correct transcriptions (a \SI{0.0}{\percent} accuracy).
This suggests that, in such cases, a more general bias design is not only beneficial but also necessary, e.g., for a document foundation model supporting various document types.

\Cref{tab:results} summarises these results.
The omitted entries (``-'') represent redundant configurations: applying a graph bias to a set record trivially reduces to the set bias, while extending a set bias to model relations yields the graph bias.

\begin{table}[t]
\centering
\begin{tabular}{c c c c c}
& & \multicolumn{3}{c}{\textbf{Inductive bias}} \\
& \textbf{} & seq & set & graph \\
\cmidrule[\heavyrulewidth]{3-5}
\multirow{3}{*}{\rotatebox[origin=c]{90}{\textbf{Record}}} & seq & \textbf{96.6 \%} & 19.4 \% & 0.0 \% \\
\cmidrule(lr){3-5} % Optional rule for better separation
& set & 12.0 \% & \textbf{74.9 \%} & - \\
\cmidrule(lr){3-5} % Optional rule for better separation
& graph & 0.0 \% & - & \textbf{74.8 \%} \\
\cmidrule[\heavyrulewidth]{3-5}
\end{tabular}
\caption{Modeling appropriate structure-specific inductive biases (bold numbers) expectedly shows a significantly better transcription accuracy than applying inappropriate biases (plain numbers).}
\label{tab:results}
\vspace{-0.75cm}
\end{table}

\subsection{Limitations and scope}

Our controlled environments rely on two primary simplifications. First, we base our custom inductive bias design on a unified base architecture, making only slight adaptations to minimize confounding architectural variables. 
Second, all experiments are conducted on synthetic, simplified data. While these simplifications serve as valid experimental constraints to substantiate our central claim, they highlight the following challenges for scaling document transcription to more realistic settings:

\textit{Inductive bias design:} 
The document-to-record task maps a specific record structure to a class of structure-specific inductive biases; however, determining the optimal design remains an open research question, particularly for non-sequential structure prediction.
While remaining-node prediction offers conceptual advantages over existing methods (see \cref{sec:related-work:relational-bias}), these must be shown to translate into empirical gains in complex, real-world settings.
Conversely, existing bipartite matching approaches, though successful in scene graph generation, must demonstrate that they can scale to graphs of higher order and size:
While we here primarily motivate structure-specific inductive biases for document transcription, the connection we establish is bidirectional; thus, document transcription itself serves as a rich and relevant testbed for improving inductive-bias design in structured-output prediction.

\textit{Data acquisition and scale:} 
Our approach assumes access to large-scale datasets containing ground-truth record representations. 
For convention-bound document types, record-aligned representations often already exist (frequently driven by the conventions themselves), making our approach directly applicable.
Examples of such representations include the Kern syntax or MusicXML for OMR, \LaTeX~for mathematical expressions, the data models generated during CAD workflows for engineering drawings, and BIM workflows for floor plans.

For domains lacking such representations, data acquisition will be costly. 
However, the document-to-record perspective offers a mitigation strategy: by grouping seemingly unrelated document types according to their shared inherent record structure, it enables cross-domain data pooling. This structural alignment enables cross-domain learning and serves as a foundational step toward developing highly capable, generalized document foundation models.

\section{Conclusion}

This paper started with the observation that documents stored as images on the one side and pictures taken in the real world on the other side have many principled differences, calling for distinct approaches for their analysis. In addition to evident dissimilarities in pixel density, which later motivated our removal of uninformative patches from model input, we highlighted fundamental distinctions in their content and meaning:
Documents are designed to convey a fixed set of information, which led to a distinct perspective: Rather than recognizing selected aspects in a document only, document recognition should be regarded as a transcription task of the full information.
An analogous transcription is impossible for complex and information-rich natural images, where no description can capture all relevant information to form a natural interface for downstream tasks.

Taking on this perspective shifts the focus of document recognition away from traditional practices -- adapting naturally misaligned computer vision approaches -- to what content must be extracted for holistic document understanding, i.e., the transcription or record of that document.
That shift to document-to-record transcription revealed several intrinsic, archetypal record structures shared across otherwise disjointed document types, such as documents containing sequential data forming linear graphs, or interlinked data forming unrestricted graphs.
These kinds of structures group document domains on a conceptual level, enabling a method of principled inductive bias design to better predict outputs following these structures. 
This is the main conceptual advance presented in this paper.

For validation, we used this method to design inductive biases for three progressively more complex record structures and learn respective document-to-record transcription models efficiently (specifically: data-efficiently) in an end-to-end framework.
In particular, based on this approach we showed a successful implementation of an end-to-end learned document recognition system for engineering drawings for the first time in the literature.
Ablation studies showed that the advance of improved transcription accuracy with less training data can indeed be attributed to the design of suitable inductive biases and not to other components of the pipeline or setup.

By framing document recognition as a domain-agnostic (but record-structure-specific) transcription task, our work opens numerous avenues for future research.
First, it enables (end-to-end) learning to understand non-sequential document types, which previously lacked a viable path to achieve similar recognition success than simpler document types.
%We present a principled inductive bias design framework--grounded in document transcription and intrinsic record structures--for document-centric deep learning methodologies.
Second, the presented method for structured inductive bias design stimulates inquiry into further structure-imposed graph types present in documents' records, such as unordered tree structures (e.g., hierarchical organizational charts) or multi-type graph compositions (for example, engineering drawing recognition can be decomposed into first recognizing the 3D shape from the 2D outlines -- an unrestricted graph, as shape primitives are interlinked --, followed by matching the annotations -- an unordered tree structure, as annotations do not annotate each other). 
Third, our work groups document types conceptually that were disjoint previously, paving the way for document foundation models trained across many document types, potentially enabling emergent capabilities in general document understanding.

Finally, beyond its theoretical implications, this perspective provides a practical rubric for end-to-end document recognition by distilling semantic information from their visual and syntactic representations. We provide a detailed summary on these practical considerations in \cref{sec:practical-scope}.

\backmatter

\bmhead{Acknowledgements}
This work has been financially supported by the Innosuisse grant 102.983 IP-ICT ``Master3D'' and the PhD support grant ``DDLearn'' from the ZHAW School of Engineering.

\bibliography{sn-article}

%\clearpage
\section*{Appendix}
\appendix

% Add appendix section to label to see in main text when appendix is referenced.
\counterwithin{figure}{section}
\counterwithin{table}{section}

\section{Proof for read function space reduction} \label{sec:app:single-function}

Here, we provide the mathematical proof that it follows from \cref{eq:convention-bound} that the read function space reduces to a single read function: $\mathcal{G} = \{ g^* \}$.
First, we restate \cref{eq:convention-bound} with the function composition definition inserted:

\begin{equation}
\forall r \in \mathcal{R}, \forall f \in \mathcal{F}, \forall g \in \mathcal{G}: g(f(r)) = r
\end{equation}

Now, we assume that there exist two functions in $\mathcal{G}$, $g_1 \in \mathcal{G}$ and $g_2 \in \mathcal{G}$ and show that they must be the same function.
Given an arbitrary write function $f_n \in \mathcal{F}$, it follows that
$g_1$ and $g_2$ must be the same function for all records written by $f_n$:

\begin{equation}
\forall r \in \mathcal{R}:\ g_1(f_n(r)) = r = g_2(f_n(r))
\end{equation}

As we chose $f_n$ arbitrarily, this holds for any $f \in \mathcal{F}$.
Thus, $g_1$ and $g_2$ are the same function for all valid documents of our domain, where valid documents are $\mathcal{D} = \{ f(r)\ |\ \forall r \in \mathcal{R}, \forall f \in \mathcal{F} \}$.
So, $g_1$ and $g_2$ are the same function across all domain-relevant inputs; we call that function $g^*$, and, therefore: $\mathcal{G} = \{ g^* \}$.

\section{Training details} \label{sec:app:training-setup}

Next to the setup described in \Cref{sec:training-setup}, we use a hardware-friendly tensor representation and batch processing to speed up training.

\textit{Tensor representation:} 
Current hardware is optimized for tensor operations. Thus, for performance considerations, we use a structure of arrays (SoA) layout:
We split the nodes of the input sequence into separate vectors: a vector of node types, a vector of discrete properties, and a vector of continuous properties. 
To ensure efficient processing and easy masking, we pad fields with varying numbers of properties so that each node occupies the same amount of space within these vectors.
With this representation, tensor operations, such as linear maps, can be run over those vectors in parallel. 
For the output prediction, we still follow the SoA layout: 
We produce a vector for the logits of all node types and a matrix for the logits of all discrete properties, padded to the maximum number of discrete properties per node type to fit a matrix structure.

\textit{Batch processing:} 
During training, we use a batch size of $32$.
Batch elements can have different sequence lengths, as records have a varying number of nodes and images a varying number of patches (we filter empty patches).
Therefore, when running the model in batch mode, we pad the shorter sequences to fit into a tensor representation.
With masking, we ensure that these padded elements do not affect the model's functionality. Specifically, we remove them for the attention mechanism and loss calculation.

\section{Record Equality} \label{sec:record-equality}

% We define record equality formally. We begin by defining node equality and edge equality.

\textit{Node equality:}
A node \(n\) consists of a type \(\text{type}(n)\) and properties \(\text{props}(n)\). 
Two nodes, \(n_1\) and \(n_2\), are equal $n_1 \eqtau n_2$ if their types match and their properties are equal (up to a precision \(\epsilon\) for continuous properties):
\begin{flalign*}
& \text{type}(n_1) = \text{type}(n_2) \land \text{props}(n_1) \eqtau \text{props}(n_2) &
\end{flalign*}
\textit{Edge equality:}
An edge consists of a type \(\text{type}(e)\) and properties \(\text{props}(e)\). Two edges, \(e_1\) and \(e_2\), are equal $e_1 \eqtau e_2$ if their types match and their properties are equal (up to a precision \(\epsilon\) for continuous properties):
\begin{flalign*}
& \text{type}(e_1) = \text{type}(e_2) \land \text{props}(e_1) \eqtau \text{props}(e_2) &
\end{flalign*}

\textit{Record equality:}
A record \(r=(N, E)\) is a property graph consisting of a set of nodes \(N\) and a set of edges \(E\), where an edge connects two nodes, a source node $n_1 = \alpha(e)$ and target node $n_2 = \beta(e)$. Two records, \(r_1=(N_1, E_1)\) and \(r_2=(N_2, E_2)\), are equal \(r_1 \eqtau r_2\) if nodes, edges and the structure are preserved. This is formalized by requiring the existence of bijective functions ensuring these properties:
\begin{flalign*}
    & \exists \text{ a bijection } f: N_1 \to N_2, & \\ 
    & \exists \text{ a bijection } g: E_1 \to E_2 \text{ s.t. } & \\
    & \ \underbrace{(\forall n \in N_1, n \eqtau f(n))}_{\text{nodes preservation}} \land \underbrace{(\forall e \in E_1, e \eqtau g(e))}_{\text{edges preservation}} \land & \\
    & \ \underbrace{(\forall e \in E_1, f(\alpha(e)) \eqtau \alpha(g(e)) \land f(\beta(e)) \eqtau \beta(g(e))}_{\text{structure preservation}} & 
\end{flalign*}

\section{Node dissimilarity} \label{sec:node-dissimilarity}

A record node $n$ has a discrete type $type(n) \in \mathcal{T}$ of a finite set of types $\mathcal{T}$ and type-specific properties $props(n)=(p_1, \hdots, p_{D+C})$. The properties consist of discrete and continuous properties: $(p_1,...,p_D)$ are $D$ discrete properties, where each $p_k \in \mathcal{V}_k$ for a finite set of values $\mathcal{V}_k$; $(p_{D+1},...,p_{D+C})$ are $C$ continuous properties, where each $p_k \in \mathbb{R}$.
A node prediction $\hat{n}$ for a node $n$ provides a predicted distribution for the type $type(\hat{n})$ and type-specific property predictions $props(\hat{n})=(\hat{p}_1,...\hat{p}_{D+C})$: $(\hat{p}_1,...,\hat{p}_D)$ are $D$ discrete property predictions, where prediction $\hat{p}_k$ is a probability distribution over the set of possible values $\mathcal{V}_k$; $(\hat{p}_{D+1},...,\hat{p}_{D+C})$ are the continuous property predictions, where prediction $\hat{p}_k \in \mathbb{R}$ is a point estimate.
The dissimilarity $l(n,\hat{n})$ between a node $n$ and a node prediction $\hat{n}$ is defined as the sum of the losses for the type and all properties. We use the cross-entropy loss, $\mathcal{L}_{CE}$, for categorical predictions and the squared error for continuous predictions. A weighting parameter $\lambda$ indicates that these discrete and continuous loss terms must be balanced in some capacity when used together: 
\begin{equation} \label{eq:node-dissimilarity}
\begin{split}
    l(n,\hat{n}) =\ & \mathcal{L}_{CE}(type(n), type(\hat{n})) \\ 
    & + \sum_{k=1}^{D} \mathcal{L}_{CE}(p_k, \hat{p}_k) \\ 
    & + \lambda \sum_{k=D+1}^{D+C} (p_k - \hat{p}_k)^2
\end{split}
\end{equation}

\section{Practical considerations} \label{sec:practical-scope}

In this section, we take a more applied viewpoint on document-as-transcription and discuss how our insights directly affect practical considerations for building document recognition systems.

\textit{Architectural implications.} The document-as-transcription perspective highlights that document recognition is a many-to-one, document-to-record task. Therefore, we can model the task as deterministic function approximation.
This avoids the target ambiguity and complex training dynamics typical of ill-posed inverse problems, where a single observation often maps to multiple valid interpretations. Furthermore, as contrasted with graph generation (see \cref{sec:related-work:relational-bias}), this deterministic mapping frees the architecture from the heavy probabilistic modeling required to learn distributions over arbitrary graphs, allowing it to optimize purely for recovering a unique target structure.

As a practical side-benefit of our document-centric focus, we exploit the inherent sparsity of convention-bound documents. By employing a vision transformer for image encoding, we can safely drop empty background patches, significantly reducing computational overhead.

\textit{Record schema design.} 
For any specific document type, the precise record schema must be defined, detailing the exact semantic entities, relations, and associated properties. 
The transcription lens guides this schema design by strictly separating semantic content from extraneous visual artifacts.
For existing record-aligned representations (e.g., Kern syntax, \LaTeX), our perspective motivates normalizing these representations to better align with the record's uniqueness and removing rendering-only information from the ground truth, as it is part of the synthesis process, not the record (cf. \cite{schmitt-koopmann_mathnet_2024}).

Crucially, not all structured formats constitute a record. 
For instance, Scalable Vector Graphics (SVGs) describe visual geometry rather than semantic topology.
A single drawn line in an SVG might represent multiple distinct semantic edges (or vice versa); thus, image vectorization \cite{su_marvel_2024} is fundamentally distinct from document transcription. 
Vectorization recovers visual paths, whereas transcription requires extracting domain-specific semantics.

\textit{Visual diversity and robustness.} 
A practical document recognition system must handle a wide array of visual styles, including pristine digital renderings, noisy scans, different fonts, or hand-drawn variations.
Within our proposed framework (see \cref{fig:framework}), such stylistic diversity is formalized as the application of distinct, valid synthesis functions.
Consequently, this diversity increases the variance of the input distribution without fundamentally altering the underlying transcription task. 
So, achieving robustness becomes primarily an engineering challenge of implementing approximate synthesis pipelines for data augmentation and adequately scaling the model's capacity (and compute).

\textit{Document and record interpretation.}
Our perspective strictly delineates document-to-record transcription from downstream interpretation. Crucially, the rasterized image and the symbolic record are informationally equivalent representations of the same underlying content. Therefore, transcription is purely a modality translation task; it does not involve interpreting this information for downstream applications, such as cost prediction for engineered parts (cf. \cite{lanfant20253d}). Meaningful document interpretation occurs subsequent to transcription, typically by applying algorithmic reasoning or graph neural networks directly to the extracted symbolic record, or potentially to the learned image embeddings.

However, we expect that semantically meaningful information is much more explicitly available in the transcribed record than the original image or any manually created partial information extraction, and that some of that semantic information will also be available in the learned internal representations of encoded document images, making them potentially useful for semantic retrieval etc.

\textit{Non-convention-bound document transcription.} 
For non-convention-bound documents, \cref{eq:convention-bound} no longer holds, as multiple interpretations about the semantics may be valid for a document.
One could still try to apply the document-as-transcription framework; however, it loses the architectural implications above, meaning that training a deterministic input-output mapping could be conceptually impossible.
Furthermore, a record schema definition may become impossible, as the document loses its intent to convey full information precisely.

While one could argue that even strict conventions might occasionally contain ambiguities that violate \cref{eq:convention-bound}, we consider these instances practical anomalies rather than systemic flaws. Such edge cases represent failures in the convention, as any ambiguity that confounds a deterministic transcription model would identically hinder human-to-human communication.

\section{Supplementary examples}

\begin{figure*}
    \centering
    \includesvg[height=0.775\textheight]{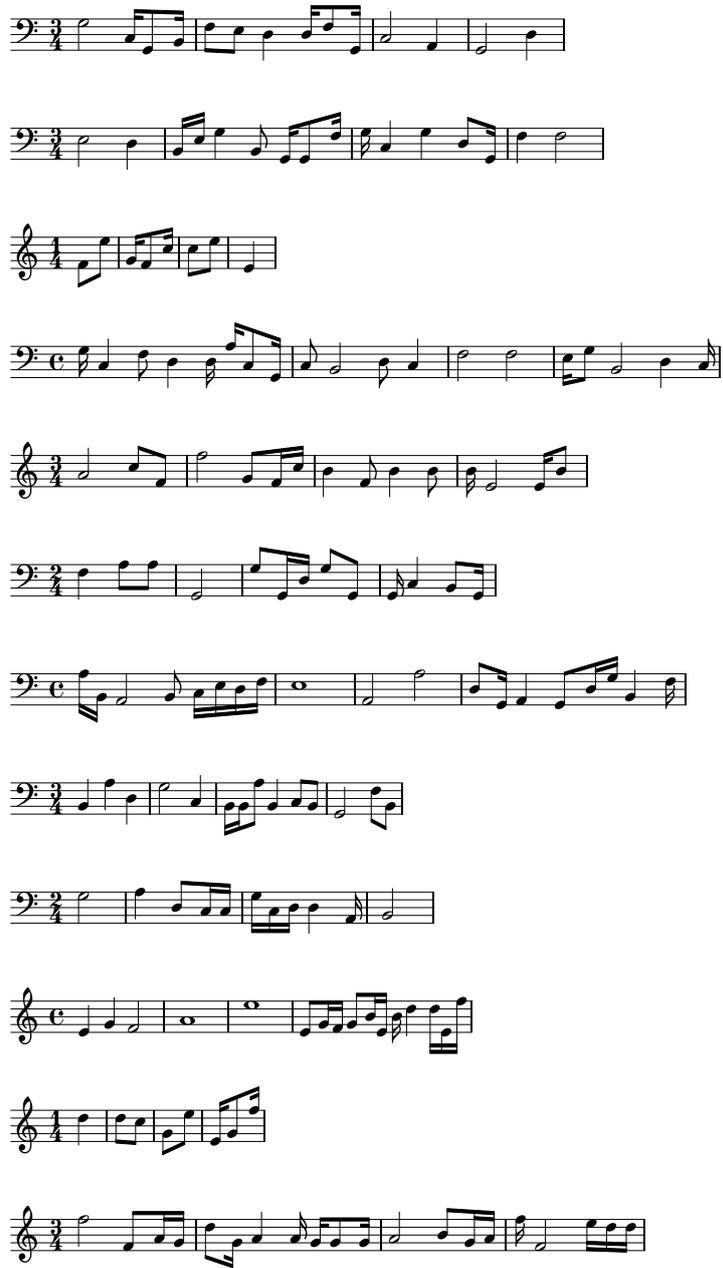}
    \caption{
    First $12$ ``sheets'' of music of the randomly generated validation set. The model exactly transcribes all of them.
    }
    \label{fig:music_data_examples}
\end{figure*}

\begin{figure*}
    \centering
    \includesvg[height=0.775\textheight]{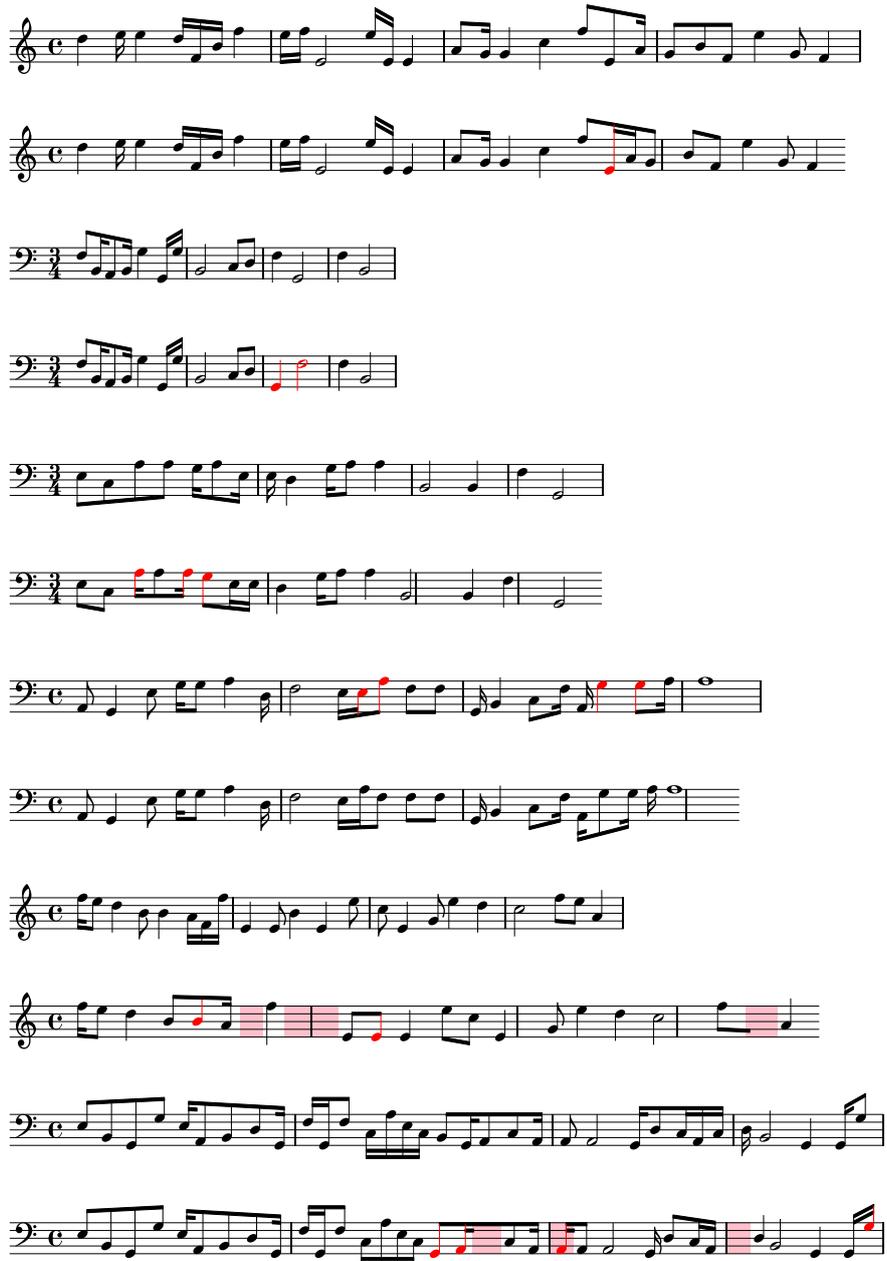}
    \caption{Six examples of incorrect transcription increasing in Levenshtein distance. First, the true sheet music is shown, followed by the corresponding prediction. In the predictions, errors are marked like this: a red note without a highlight means at least one music note property was wrong (replace), an empty red highlight means the prediction had a music note missing (delete), a red note in red highlight means the prediction had an extra music note (insert).}
    \label{fig:music_faulty_examples}
\end{figure*}

\begin{figure*}
\vspace{-2cm}
\begin{subfigure}{\textwidth}
\begin{subfigure}{0.48\textwidth}
\begin{lstlisting}[language=LilyPond,escapechar=!]]
!\color{magenta}{Line(start=(0.18, 0.65), end=(0.48, 0.65))}!
!\color{green}{Line(start=(0.18, 0.44), end=(0.18, 0.65))}!
!\color{violet}{Line(start=(0.48, 0.85), end=(0.67, 0.85))}!
!\color{brown}{Line(start=(0.67, 0.85), end=(0.84, 0.6))}!
!\color{blue}{Line(start=(0.48, 0.24), end=(0.67, 0.24))}!
!\color{orange}{Line(start=(0.18, 0.44), end=(0.48, 0.44))}!
!\color{pink}{Line(start=(0.67, 0.24), end=(0.84, 0.49))}!
!\color{black}{Line(start=(0.84, 0.49), end=(0.84, 0.6))}!
!\color{cyan}{Line(start=(0.48, 0.24), end=(0.48, 0.85))}!
!\color{gray}{Line(start=(0.67, 0.24), end=(0.67, 0.85))}!
\end{lstlisting}
\end{subfigure}
\hfill
\begin{subfigure}{0.48\textwidth}
    \hspace{-0.09\textwidth}
    \includegraphics[width=0.75\linewidth]{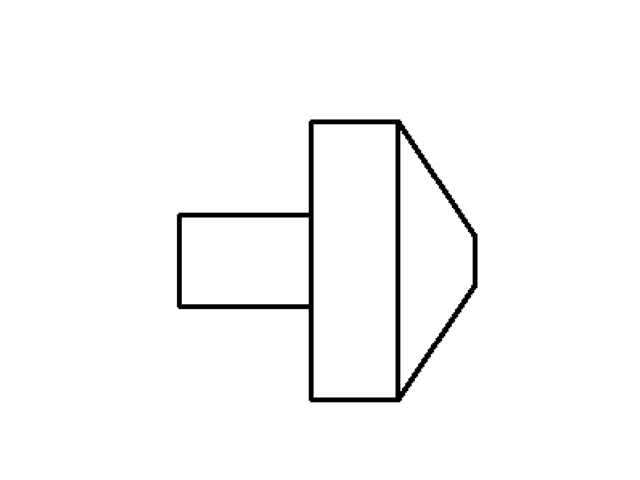}
\end{subfigure}

% Add arrow between the subfigures
\begin{tikzpicture}[overlay]
    \draw[-stealth, shorten >=3pt] (7.75, 3) -- node [midway, above,xshift=-2pt] {\scriptsize Synthesis}  (9, 3);
    \draw[-stealth, shorten >=3pt] (9.5, 1) -- node [midway, left] {\scriptsize Transcription} (7.75, -1.25);
    \draw[-stealth, shorten >=3pt] (7.75, -1.75) -- node [midway, below] {\scriptsize Synthesis}  (9, -1.75);
    \draw[-stealth, shorten >=3pt] (11.5, 1) -- (12.5, 0.5);
    \draw[-stealth, shorten >=3pt] (11.5, -1) -- (12.5, -0.5);
\end{tikzpicture}

\begin{subfigure}{0.48\textwidth}
\begin{lstlisting}[language=LilyPond,escapechar=!]]
!\color{blue}{Line(start=(0.48, 0.24), end=(0.67, 0.24))}!
!\color{black}{Line(start=(0.84, 0.49), end=(0.84, 0.6))}!
!\color{green}{Line(start=(0.19, 0.44), end=(0.19, 0.65))}!
!\color{violet}{Line(start=(0.48, 0.85), end=(0.67, 0.85))}!
!\color{orange}{Line(start=(0.19, 0.44), end=(0.48, 0.44))}!
!\color{gray}{Line(start=(0.67, 0.24), end=(0.67, 0.85))}!
!\color{pink}{Line(start=(0.67, 0.24), end=(0.84, 0.49))}!
!\color{brown}{Line(start=(0.67, 0.85), end=(0.84, 0.6))}!
!\color{cyan}{Line(start=(0.48, 0.24), end=(0.48, 0.86))}!
!\color{magenta}{Line(start=(0.19, 0.65), end=(0.48, 0.65))}!
\end{lstlisting}
\end{subfigure}
\hfill
\begin{subfigure}{0.48\textwidth}
    \hspace{-0.09\textwidth}
    \vspace{-0.1\textwidth}
    \includegraphics[width=0.75\linewidth]{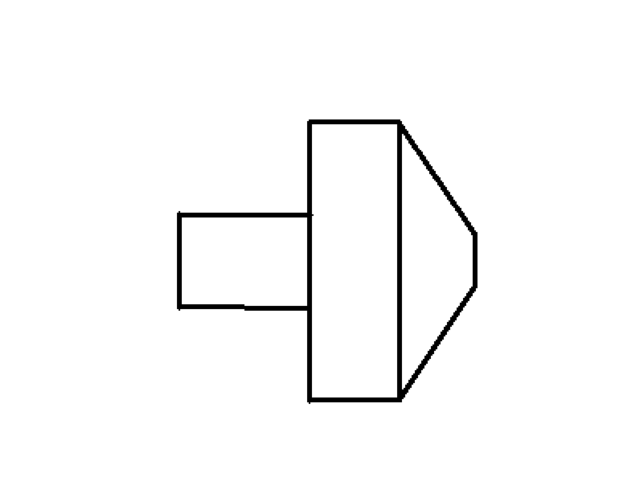}
\end{subfigure}
\begin{tikzpicture}[overlay]
\begin{subfigure}{0.48\textwidth}
    \hspace{-0.65\textwidth}
    \vspace{0.275\textwidth}
    \includegraphics[width=0.75\linewidth]{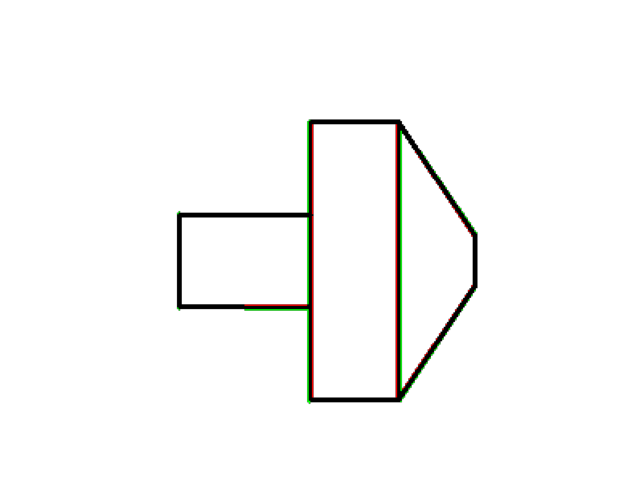}
\end{subfigure}
\end{tikzpicture}
\caption{Correct transcription.}
\end{subfigure}

\vspace{-0.5cm}

\begin{subfigure}{\textwidth}
\begin{subfigure}{0.48\textwidth}
\begin{lstlisting}[language=LilyPond,escapechar=!]]
!\color{magenta}{Line(start=(0.48, 0.18), end=(0.71, 0.18))}!
!\color{green}{Line(start=(0.40, 0.67), end=(0.79, 0.67))}!
!\color{violet}{Line(start=(0.48, 0.18), end=(0.48, 0.51))}!
!\color{brown}{Line(start=(0.71, 0.18), end=(0.71, 0.51))}!
!\color{blue}{Line(start=(0.46, 0.81), end=(0.74, 0.81))}!
!\color{orange}{Line(start=(0.40, 0.51), end=(0.40, 0.67))}!
!\color{pink}{Line(start=(0.74, 0.67), end=(0.74, 0.81))}!
!\color{black}{Line(start=(0.46, 0.67), end=(0.46, 0.81))}!
!\color{cyan}{Line(start=(0.79, 0.51), end=(0.79, 0.67))}!
!\color{gray}{Line(start=(0.40, 0.51), end=(0.79, 0.51))}!
\end{lstlisting}
\end{subfigure}
\hfill
\begin{subfigure}{0.48\textwidth}
    \hspace{-0.2\textwidth}
    \includegraphics[width=0.9\linewidth]{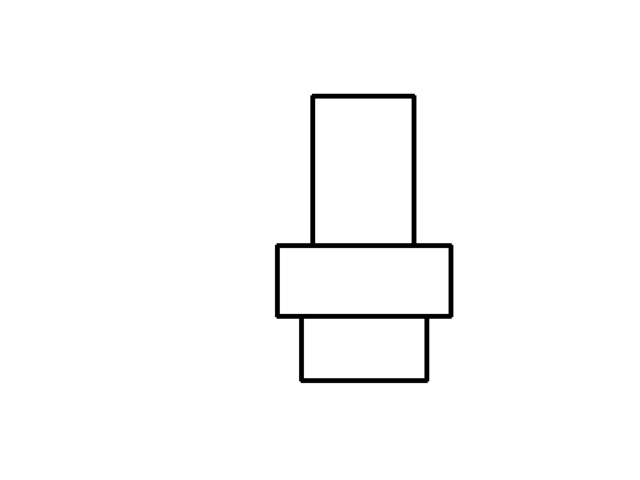}
\end{subfigure}
\hfill

% Add arrow between the subfigures
\begin{tikzpicture}[overlay]
    \draw[-stealth, shorten >=3pt] (7.75, 3) -- node [midway, above] {\scriptsize Synthesis} (9, 3);
    \draw[-stealth, shorten >=3pt] (9.5, 1) -- node [midway, left] {\scriptsize Transcription} (7.75, -1.25);
    \draw[-stealth, shorten >=3pt] (7.75, -1.75) -- node [midway, below] {\scriptsize Synthesis}  (9, -1.75);
    \draw[-stealth, shorten >=3pt] (11.5, 1) -- (12.5, 0.5);
    \draw[-stealth, shorten >=3pt] (11.5, -1) -- (12.5, -0.5);
\end{tikzpicture}

\begin{subfigure}{0.48\textwidth}
\begin{lstlisting}[language=LilyPond,escapechar=!]]
!\color{blue}{Line(start=(0.46, 0.81), end=(0.74, 0.81))}!
!\color{cyan}{Line(start=(0.79, 0.51), end=(0.79, 0.67))}!
!\color{orange}{Line(start=(0.40, 0.51), end=(0.40, 0.67))}!
!\color{gray}{Line(start=(0.40, 0.51), end=(0.79, 0.51))}!
!\color{pink}{Line(start=(0.74, 0.67), end=(0.74, 0.81))}!
!\color{green}{Line(start=(0.41, 0.67), end=(0.79, 0.67))}!
!\color{black}{Line(start=(0.46, 0.67), end=(0.46, 0.81))}!
!\color{violet}{Line(start=(0.48, 0.18), end=(0.48, 0.51))}!
!\color{brown}{Line(start=(0.71, 0.18),}! !\color{red}{\underline{end=(0.71, 0.18)}}!!\color{brown}{)}!
!\color{magenta}{Line(start=(0.48, 0.18), end=(0.71, 0.18))}!
\end{lstlisting}
\end{subfigure}
\hfill
\begin{subfigure}{0.48\textwidth}
    \hspace{-0.2\textwidth}
    \vspace{-0.1\textwidth}
    \includegraphics[width=0.9\linewidth]{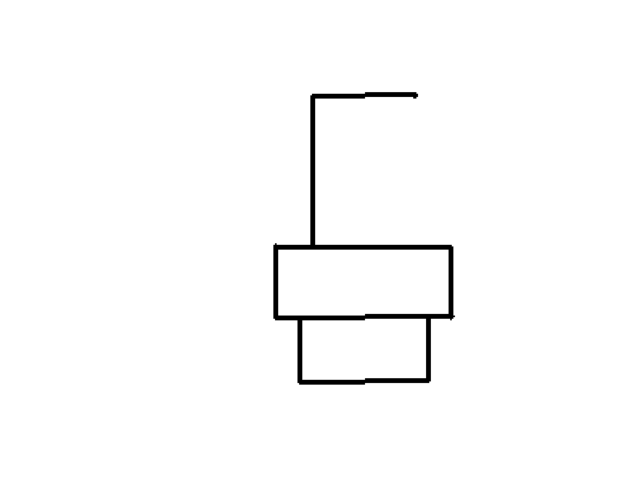}
\end{subfigure}
\begin{tikzpicture}[overlay]
\begin{subfigure}{0.48\textwidth}
    \hspace{-0.75\textwidth}
    \vspace{0.275\textwidth}
    \includegraphics[width=0.9\linewidth]{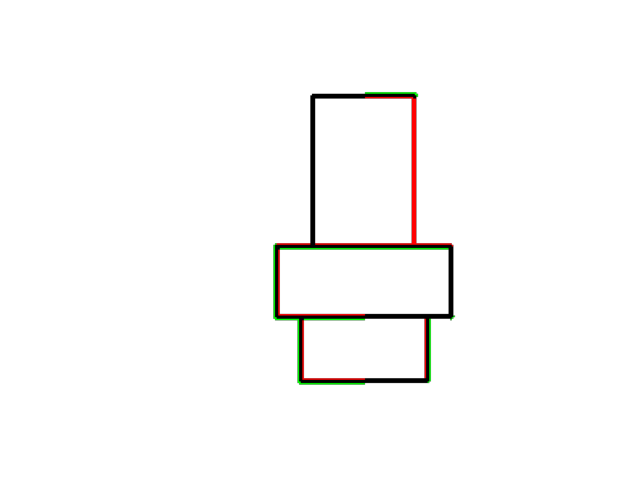}
\end{subfigure}
\end{tikzpicture}

\caption{Incorrect transcription (errors highlighted with red underlined text).}
\end{subfigure}

\caption{Example of a correct (a) and an incorrect (b) transcription of shape drawings.
In both cases, we start with the true record (top left) and render it (top right). Then, the model predicts a record (bottom left), which we render for easier comparison (bottom right). Finally, we render the prediction in green on top of the true shape in red, and perfect overlapping matches in black (middle right).}
\label{fig:technical_drawing_prediction_example}
\vspace{-2cm}
\end{figure*}

\begin{figure*}[p]
    \centering
    \includegraphics[height=0.84\textheight]{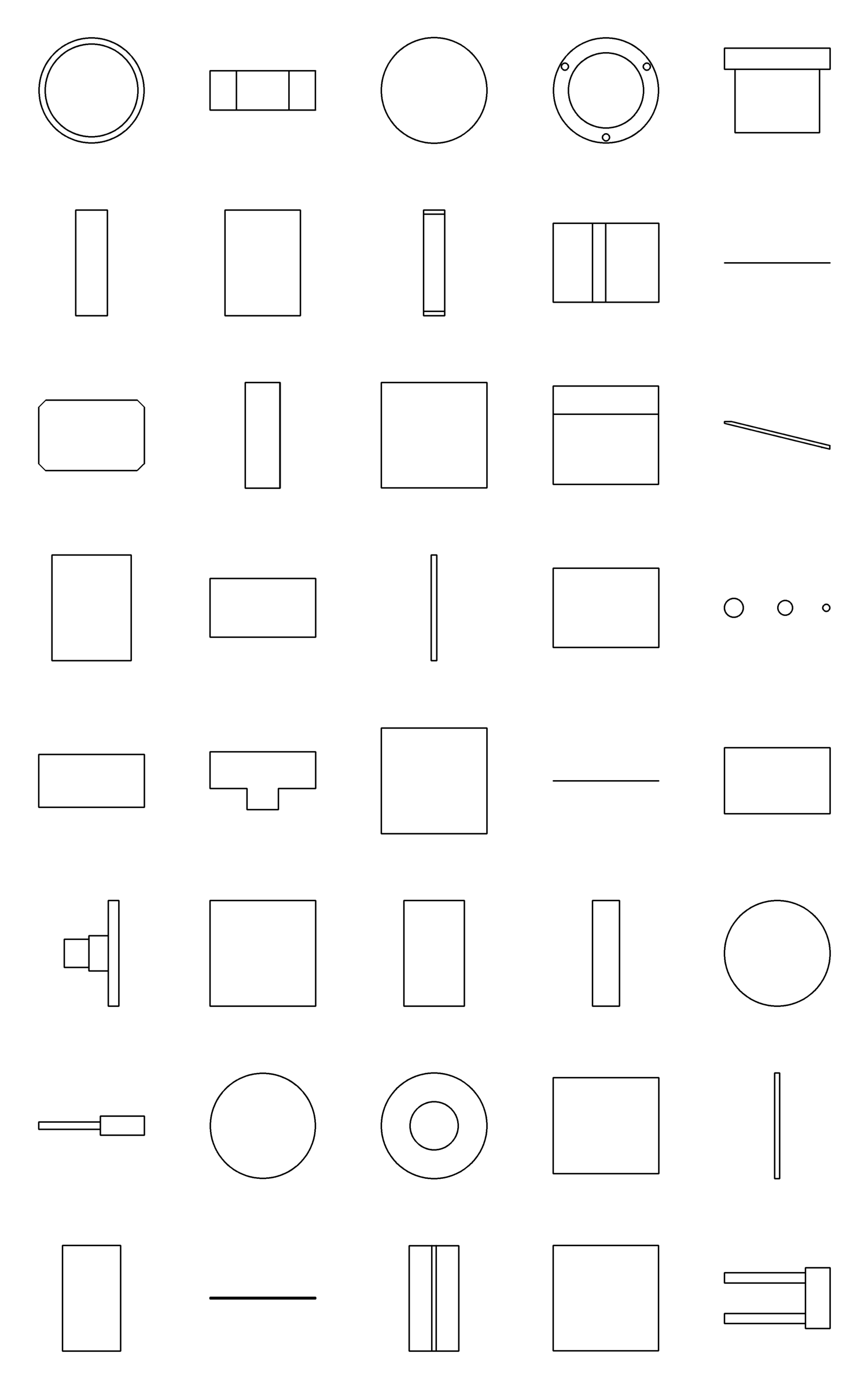}
    \caption{Examples of $40$ engineering drawings. One cell represents one example.}
    \label{fig:abc_data_examples}
\end{figure*}

\begin{figure*}[p]
    \centering
    \includegraphics[height=0.84\textheight]{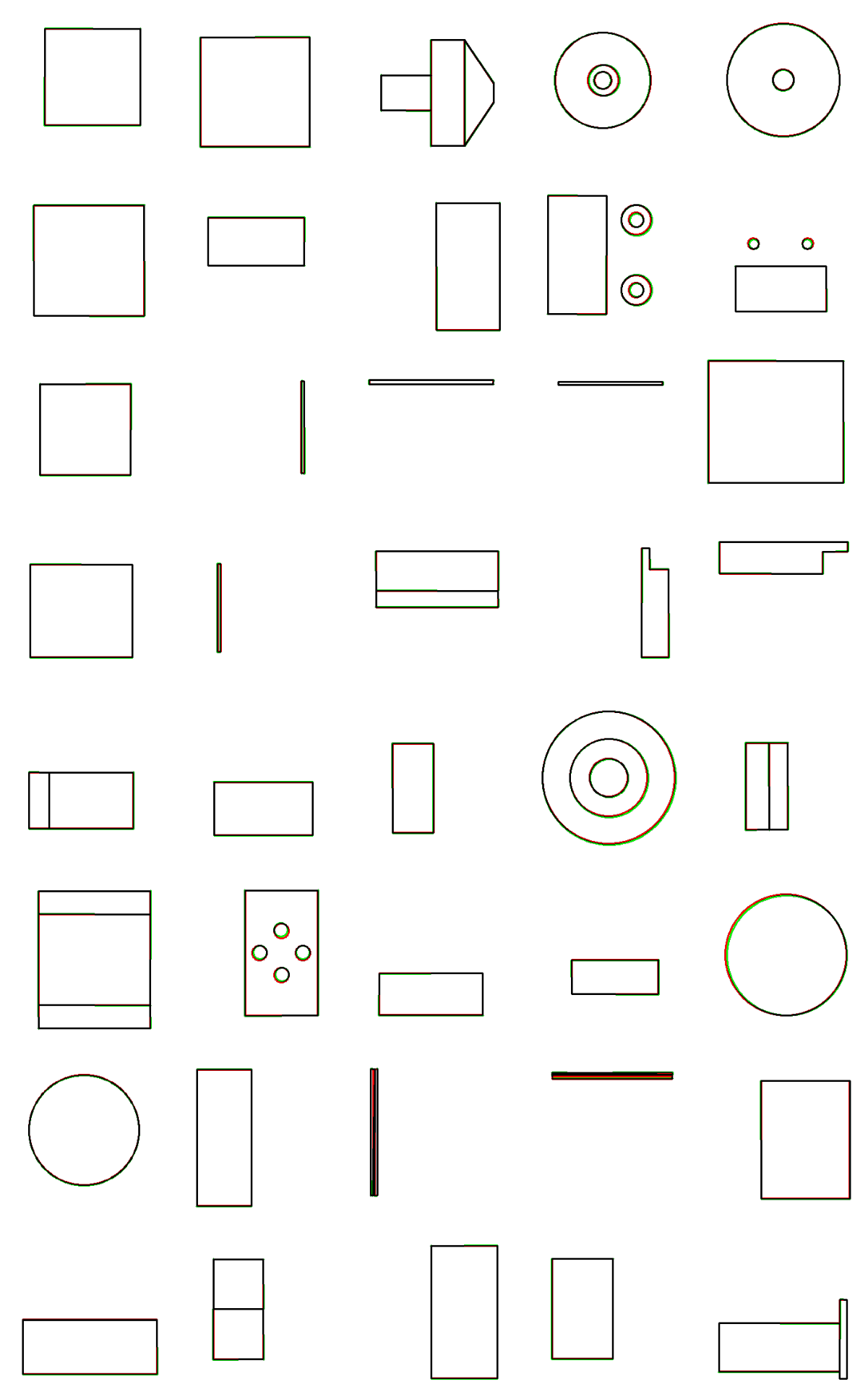}
    \caption{First $40$ examples of correct transcriptions in the validation set. The ground truth
    is drawn in red, the prediction in green. If overlapping exactly, the line is drawn in black.}
    \label{fig:technical_drawings_success_examples}
\end{figure*}

\begin{figure*}[p]
    \centering
    \includegraphics[height=0.84\textheight]{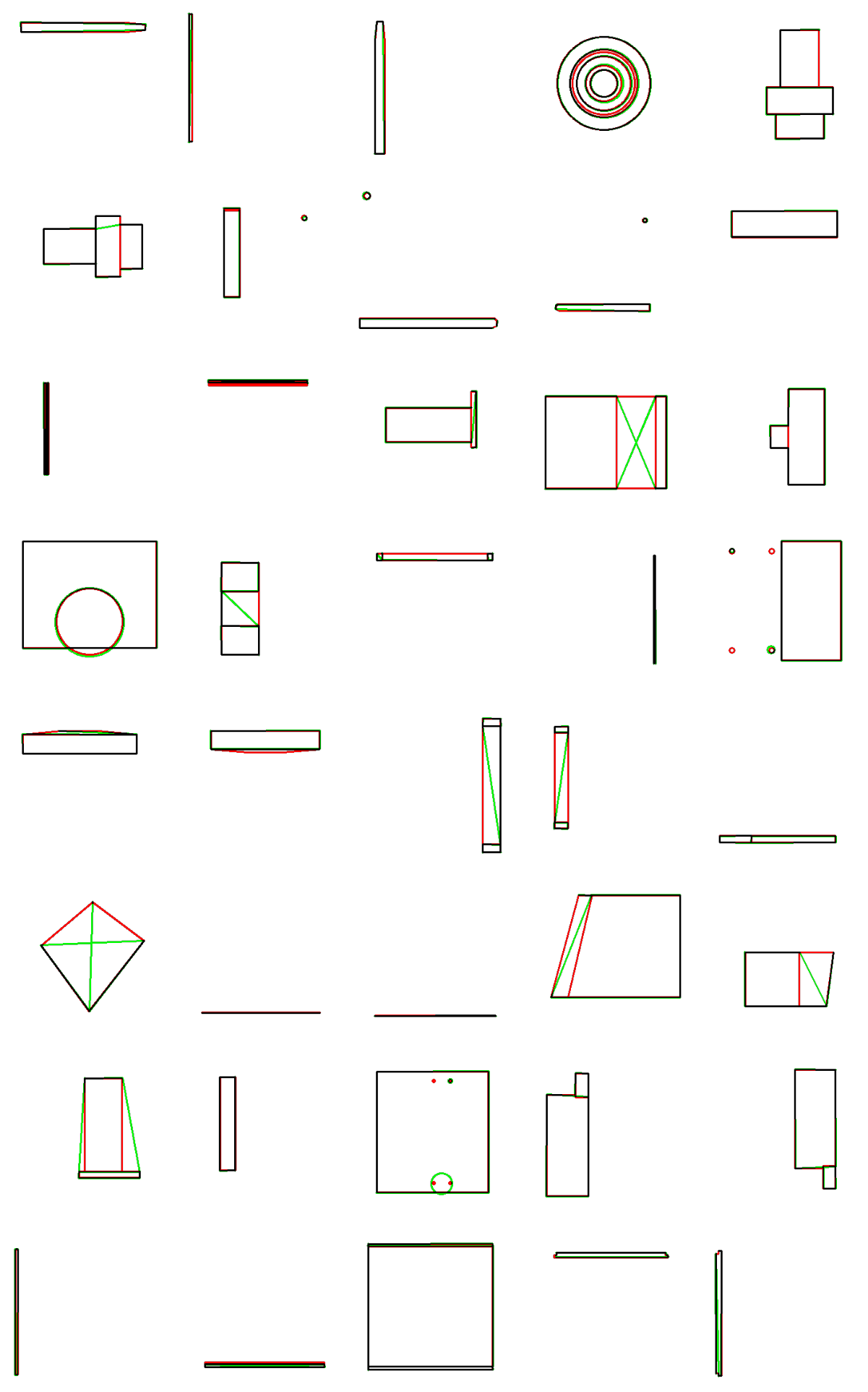}
    \caption{First $40$ examples of failed transcriptions in the validation set. The ground truth
    is drawn in red, the prediction in green. If overlapping exactly, the line is drawn in black.}
    \label{fig:technical_drawings_faulty_examples}
\end{figure*}

\begin{figure*}[p]
    \centering
    \includegraphics[width=0.8\textwidth]{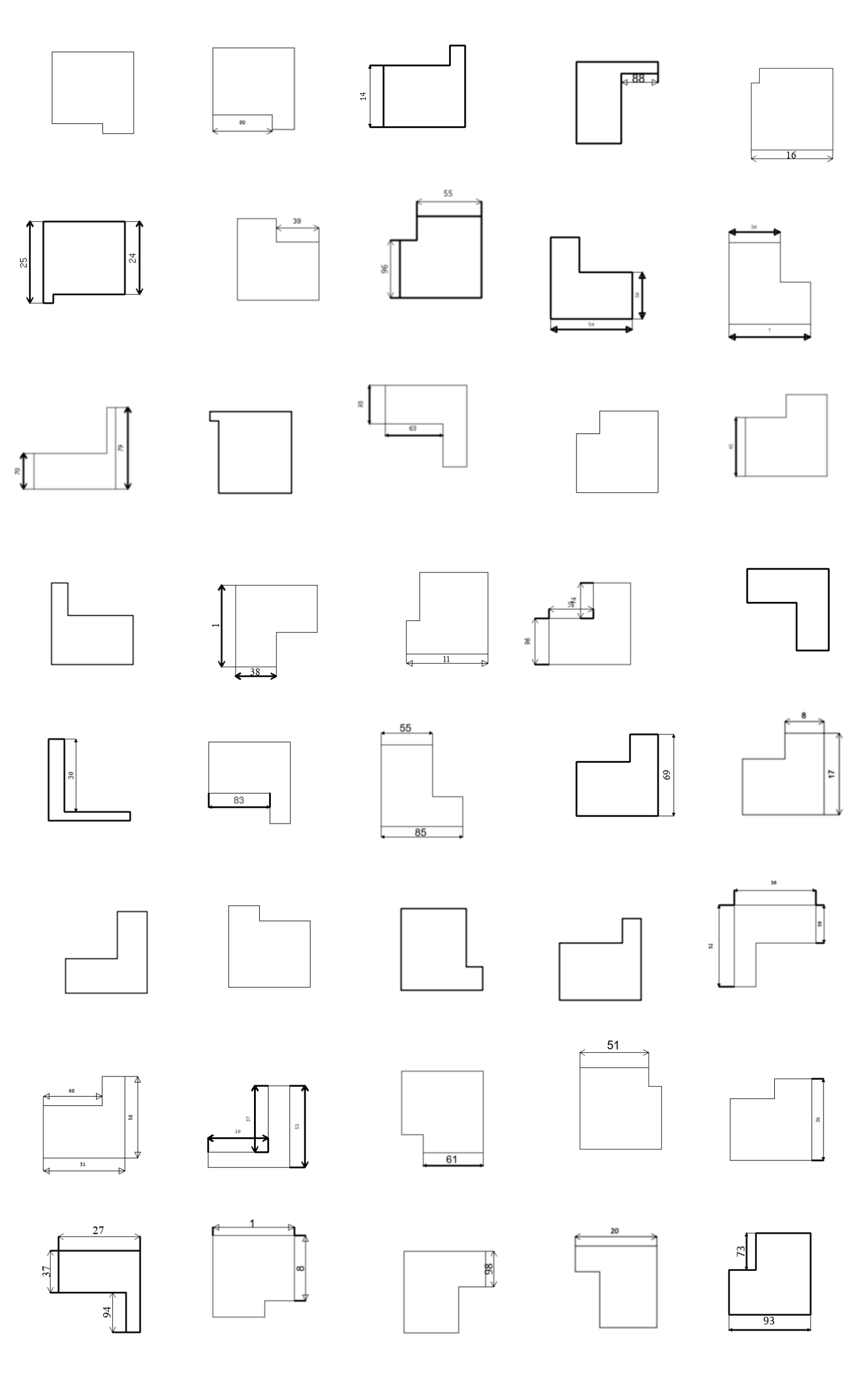}
    \caption{Examples of synthetically generated engineering drawings.}
    \label{fig:l_shape_examples}
\end{figure*}

\Cref{fig:music_pred_example} shows the synthesis and transcription process for sheet music. 
\Cref{fig:music_data_examples} shows examples of successfully transcribed sheet music. 
\Cref{fig:music_faulty_examples} shows examples of incorrectly transcribed sheet music.
\Cref{fig:technical_drawing_prediction_example} shows the synthesis and transcription process for shape drawings. 
\Cref{fig:abc_data_examples} shows examples of shape drawings. 
\Cref{fig:technical_drawings_success_examples} shows examples of correctly transcribed shape drawings.
\Cref{fig:technical_drawings_faulty_examples} shows examples of incorrectly transcribed shape drawings.
\Cref{fig:l_shape_examples} shows examples of simplified engineering drawings.

\begin{figure*}[b]
\begin{subfigure}{\textwidth}
\begin{subfigure}{0.48\textwidth}
\begin{lstlisting}[language=LilyPond]
\clef treble \time 3/4 f''2 f'8 a'16 g'16 d''8 g'16 a'4 a'16 g'16 g'8 g'16 a'2 b'8 g'16 a'16 f''16 f'2 e''16 d''16 d''16 
\end{lstlisting}
\end{subfigure}
\hfill
\begin{subfigure}{0.48\textwidth}
    \includesvg[width=\linewidth]{img/appendix/music_success_example.svg}
\end{subfigure}

% Add arrow between the subfigures
\begin{tikzpicture}[overlay]
    \draw[-stealth, shorten >=3pt] (7.75, 1.0) -- node [midway, above] {\scriptsize Synthesis} (8.25, 1.0);
    \draw[-stealth, shorten >=3pt] (8.25, 0.25) -- node [midway, right,xshift=4pt] {\scriptsize Transcription} (7.75, -0.25);
    \draw[-stealth, shorten >=3pt] (7.75, -1.0) -- node [midway, below] {\scriptsize Synthesis} (8.25, -1.0);
\end{tikzpicture}

\vspace*{0.1cm}

\begin{subfigure}{0.48\textwidth}
\begin{lstlisting}[language=LilyPondPred]
\clef treble \time 3/4 f''2 f'8 a'16 g'16 d''8 g'16 a'4 a'16 g'16 g'8 g'16 a'2 b'8 g'16 a'16 f''16 f'2 e''16 d''16 d''16 
\end{lstlisting}
\end{subfigure}
\hfill
\begin{subfigure}{0.48\textwidth}
    \includesvg[width=\linewidth]{img/appendix/music_success_example.svg}
\end{subfigure}
\vspace*{0.3cm}
\caption{Correct transcription.}
\end{subfigure}

\vspace*{0.5cm}

\begin{subfigure}{\textwidth}
\begin{subfigure}{0.48\textwidth}
\begin{lstlisting}[language=LilyPond,escapechar=!]
\clef treble \time 3/4 b'8 e''16 e'8 c''8 g'16 e'16 d''16 d''8 d''4 b'8 g'16 f''16 c''!\color{red}{16}! c''!\color{red}{8}! e'16 e'8 a'16 b'16 a'2 f''2 b'8 f'8 
\end{lstlisting}
\end{subfigure}
\hfill
\begin{subfigure}{0.48\textwidth}
    \includesvg[width=\linewidth]{img/appendix/music_faulty_example_truth.svg}
\end{subfigure}

% Add arrow between the subfigures
\begin{tikzpicture}[overlay]
    \draw[-stealth, shorten >=3pt] (7.75, 1.0) -- node [midway, above] {\scriptsize Synthesis} (8.25, 1.0);
    \draw[-stealth, shorten >=3pt] (8.25, 0.25) --  node [midway, right,xshift=4pt] {\scriptsize Transcription} (7.75, -0.25);
    \draw[-stealth, shorten >=3pt] (7.75, -1.0) -- node [midway, below] {\scriptsize Synthesis} (8.25, -1.0);
\end{tikzpicture}

\vspace*{0.2cm}

\begin{subfigure}{0.48\textwidth}
\begin{lstlisting}[language=LilyPondPred,escapechar=!]
\clef treble \time 3/4 b'8 e''16 e'8 c''8 g'16 e'16 d''16 d''8 d''4 b'8 g'16 f''16 c''!\color{red}{\underline{8}}! c''!\color{red}{\underline{16}}! e'16 e'8 a'16 b'16 a'2 f''2 b'8 f'8 
\end{lstlisting}
\end{subfigure}
\hfill
\begin{subfigure}{0.48\textwidth}
    \centering
    \includesvg[width=\linewidth]{img/appendix/music_faulty_example_pred.svg}
    \label{fig:enter-label}
\end{subfigure}
\vspace*{0.2cm}
\caption{Incorrect transcription (errors highlighted with red underlined text).}
\end{subfigure}
\caption{
Example of a correct (a) and an incorrect transcription (b).
Both start with the true record (top left, written in LilyPond syntax) and rendering (top right). Then, the model predicts a record (bottom left, written in LilyPond syntax), which is rendered for visual comparison (bottom right).
} \label{fig:music_pred_example}
\end{figure*}

%\subsection{Engineering drawings correct transcription examples}

%\subsection{Engineering drawings incorrect transcription examples}

\end{document}